\documentclass{article}

\usepackage{color,xcolor}
\usepackage{epsfig}
\usepackage{graphicx}

\usepackage{adjustbox}
\usepackage{array}
\usepackage{booktabs}
\usepackage{colortbl}
\usepackage{float,wrapfig}
\usepackage{hhline}
\usepackage{multirow}
\usepackage[font={small}]{caption}
\usepackage{natbib}

\usepackage{amsmath,amsfonts,amsthm,amssymb}
\usepackage{bm}
\usepackage{nicefrac}
\usepackage{microtype}
\usepackage{mathtools}

\usepackage{changepage}
\usepackage{extramarks}
\usepackage{fancyhdr}
\usepackage{lastpage}
\usepackage{setspace}
\usepackage{soul}
\usepackage{xspace}

\usepackage{url}

\usepackage{algorithmic}
\usepackage{algorithm}
\usepackage{enumitem}
\usepackage{titlesec}
\usepackage{bbm}
\usepackage{soul}
\usepackage{makecell}
\usepackage{microtype}
\usepackage{graphicx}
\usepackage{subfigure}
\usepackage{booktabs} 
\newcommand{\model}{NIIP\xspace}
\newcommand{\modeltit}{Neural Interaction Inference with Potentials\xspace}
\newcommand{\myparagraph}[1]{\vspace{-8pt}\paragraph{#1}}
\newcommand{\myitem}{\vspace{-3pt}\item}

\newcommand{\sect}[1]{Section~\ref{#1}}

\newcommand{\eqn}[1]{Equation~\ref{#1}}
\newcommand{\fig}[1]{Figure~\ref{#1}}

\usepackage{amsmath,amsfonts,bm}

\def\eqref#1{equation~\ref{#1}}

\def\1{\bm{1}}

\def\vx{{\bm{x}}}

\def\vz{{\bm{z}}}

\DeclareMathAlphabet{\mathsfit}{\encodingdefault}{\sfdefault}{m}{sl}
\SetMathAlphabet{\mathsfit}{bold}{\encodingdefault}{\sfdefault}{bx}{n}

\newcommand{\E}{\mathbb{E}}

\usepackage[pagebackref=true,breaklinks=true,colorlinks,citecolor=brown]{hyperref}

\usepackage[accepted]{icml2023_new}

\usepackage{amsmath}
\usepackage{amssymb}
\usepackage{mathtools}
\usepackage{amsthm}

\usepackage[capitalize,noabbrev]{cleveref}

\theoremstyle{plain}

\theoremstyle{definition}

\theoremstyle{remark}

\usepackage[textsize=tiny]{todonotes}

\newcommand{\real}{\mathbb{R}}

\newcommand{\V}[1]{{\mathbf{#1}}} 
\usepackage{graphicx}
\usepackage{enumitem}
\usepackage{comment}
\usepackage{enumitem}
\usepackage{wrapfig}
\usepackage{xcolor}
\usepackage{booktabs}
\usepackage{makecell}

\usepackage{subfloat}

\hypersetup{colorlinks=true,linkcolor=red,citecolor=brown,urlcolor=blue}

\icmltitlerunning{Inferring Relational Potentials in Interacting Systems}

\begin{document}

\twocolumn[
\icmltitle{Inferring Relational Potentials in Interacting Systems}

\icmlsetsymbol{equal}{*}

\begin{icmlauthorlist}
\icmlauthor{Armand {Comas Massagu\'{e}}}{neu}
\icmlauthor{Yilun Du}{mit}
\icmlauthor{Christian Fernandez}{neu}
\icmlauthor{Sandesh Ghimire}{neu}
\icmlauthor{Mario Sznaier}{neu}
\icmlauthor{Joshua B. Tenenbaum}{mit}
\icmlauthor{Octavia Camps}{neu}
\end{icmlauthorlist}

\icmlaffiliation{neu}{Northeastern University}
\icmlaffiliation{mit}{Massachusetts Institute of Technology}

\icmlcorrespondingauthor{Armand Comas}{comasmassague.a@northeastern.edu}

\icmlkeywords{Machine Learning, ICML}

\vskip 0.3in
]

\printAffiliationsAndNotice{} 

\begin{abstract}
Systems consisting of interacting agents are prevalent in the world, ranging from dynamical systems in physics to complex biological networks. To build systems which can interact robustly in the real world, it is thus important to be able to infer the precise interactions governing such systems.  Existing approaches typically discover such interactions by explicitly modeling the feed-forward dynamics of the trajectories. In this work, we propose Neural Interaction Inference with Potentials (\model) as an alternative approach to discover such interactions that enables greater flexibility in trajectory modeling: it discovers a set of relational potentials, represented as energy functions, which when minimized reconstruct the original trajectory. 
\model assigns low energy to the subset of trajectories which respect the relational constraints observed. 
We illustrate that with these representations \model displays unique capabilities in test-time. First, it allows trajectory manipulation, such as interchanging interaction types across separately trained models, as well as trajectory forecasting. Additionally, it allows adding external hand-crafted potentials at test-time. Finally, \model enables the detection of out-of-distribution samples and anomalies without explicit training. 
Website:  \url{https://energy-based-model.github.io/interaction-potentials}.

\end{abstract}

\section{Introduction}

\begin{figure}[t]
\centering
\includegraphics[trim={20 670 1300 0}, width=0.45\textwidth]{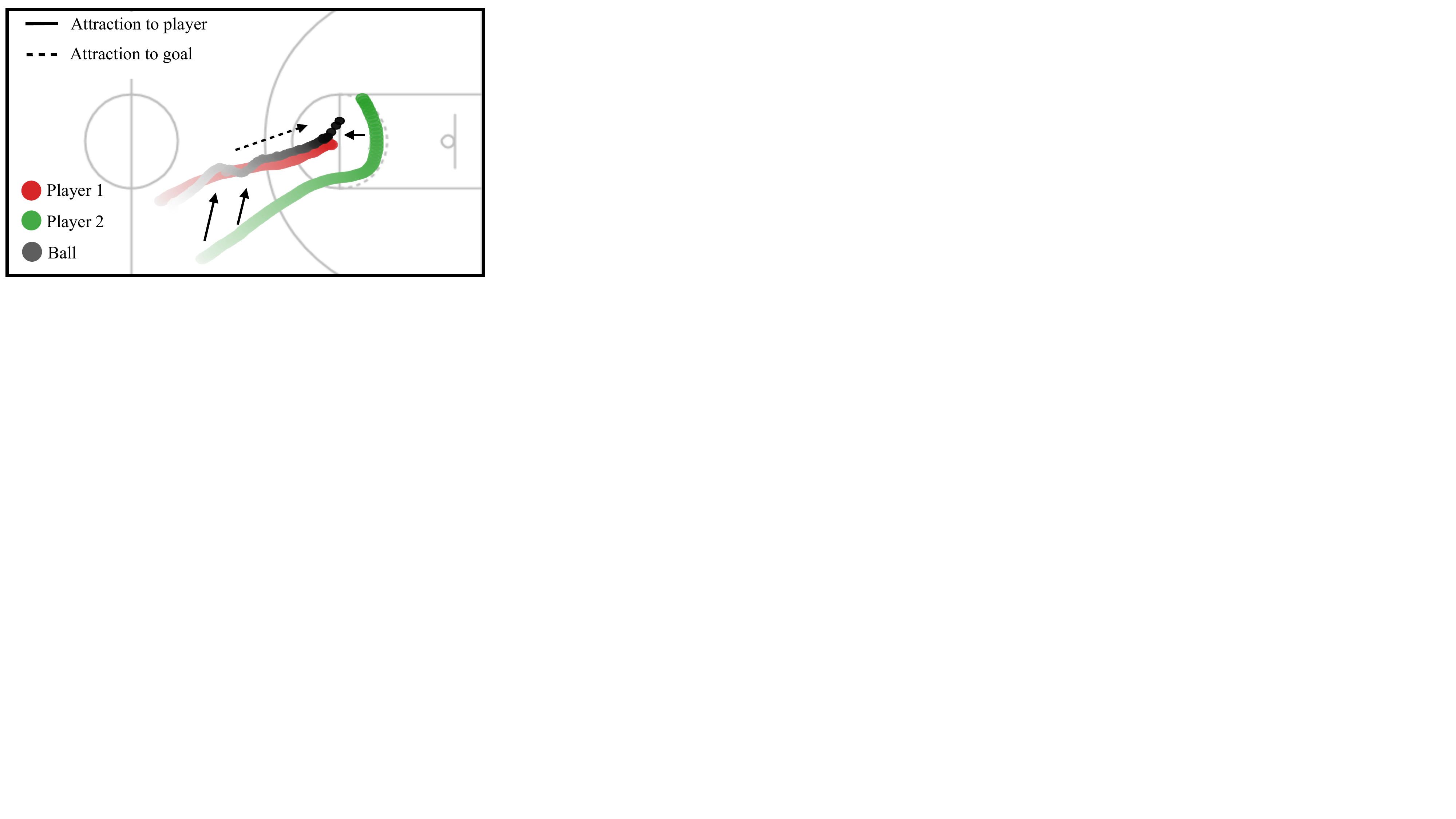}
\vspace{-20pt}
  \caption{\small \textbf{Interactions between NBA players.} Complex dynamics, such as the player trajectories in the NBA, may be explained using a simple set of interactions. In this setting, one player aims to block the other one from scoring.}\label{fig:teaser}
  \vspace{-15pt}
\end{figure}

Dynamical systems are ubiquitous in both nature and everyday life. Such systems emerge naturally in scientific settings such as chemical pathways and particle dynamics as well as everyday settings such as in sports teams or social events. Such dynamical systems may be decomposed as a set of different interacting components, where the interactions among them lead to complex dynamics. Modeling the dynamics of such systems is hard: often times we only have access to example trajectories, without knowledge of the underlying interactions or the dynamics that govern them.

Consider the scenario given in Figure \ref{fig:teaser}, consisting of a two NBA players playing a basketball game (the rest of the teams are ommited for clarity). While the motion of individual players may appear stochastic in nature, each player aims to score the basket on the opposite team's side of the court. Thus, we may utilize sets of interactions to explain their behaviors -- One or more offensive players move towards the goal, while a defensive player moves to intercept them and prevent them from scoring. By applying our underlying knowledge of these interactions between players, we may  forecast the future dynamics of the basketball game significantly more accurately. 

Most works modeling such complex dynamics do not explicitly disentangle individual interactions between objects.  Instead, they rely on a learned network to implicitly disentangle them \citep{Battaglia2016InteractionNF, Gilmer2017NeuralMP, Steenkiste2018RelationalNE}. In contrast, \cite{Kipf2018NeuralRI} propose Neural Relation Inference (NRI), which learns a structured set of explicit interaction models between objects and show how such explicit interaction modeling  enables more effective downstream predictions.

In this work, we argue that we should instead model and disentangle interactions between objects as a set of learned interaction potentials, with dynamical prediction corresponding to a potential satisfaction problem. To this end, we propose \modeltit (\model), where we encode each of these potentials as an energy function \citep{Du2021UnsupervisedLO}.  

In physics, a potential is the energy held by an object because of its position relative to other objects. To predict future dynamics with \model, we solve a potential minimization problem, where we optimize for a trajectory prediction which minimizes our predicted energy. In different experiments, we illustrate how our potential-based decomposition of interactions provides unique benefits over prior learned approaches for decomposing dynamics. 

First, we show that the learned potentials are disentangled and composable, enabling us to interchange interaction types across separate models, trained with different datasets. We also illustrate that such a decomposition enables us to add flexible test-time potentials to incorporate new changes in the environment. 
We further show that \model can detect anomalies (out-of-distribution relation types) without a dedicated training. Finally, we compare our model to existing approaches for trajectory forecasting in both synthetic and real settings, illustrating that \model performs well in of mid- and long- term prediction.

In summary, in this work, we contribute the following: \textbf{(i).} We propose \modeltit (\model) a novel method that discovers, in an unsupervised manner, the underlying interactions between particles in a system as a set of energy functions or potentials. \textbf{(ii).} We illustrate how such a potential-based decomposition of interactions offers unique properties related to composability, such as interchanging interaction types across separate models or adding new hand-crafted potentials at test-time. \textbf{(iii).} We further show how \model can detect out-of-distribution samples by design, without further training.
And \textbf{(iv).} we illustrate how such a potential decomposition of interactions enables accurate mid- and long-horizon trajectory prediction performance, often surpassing existing methods.


\vspace{-5pt}
\section{Potentials as Energy Based Models} \label{sec:background}
We will consider \textit{potentials} as specifying a set $X$ of trajectories $\vx \in \mathbb{R}^{T \times D}$ which possess \textcolor{red}{an} underlying property we desire. In section \sect{sect:ebm}, we discuss how we can represent potentials on trajectories using an EBM. We further discuss how we may compose multiple potentials together as EBMs in \sect{sect:compose_constraint}.

\vspace{-3pt}
\subsection{Energy-based Models}
\label{sect:ebm}
\vspace{5pt}
\myparagraph{Definition.} An Energy-Based Model (EBM) is defined probabilistically using the Boltzmann distribution $p_\theta(\vx) = \frac{\exp(-E_\theta(\vx))}{Z(\theta)}$, with an underlying partition function $Z(\theta) = \int \exp(-E_\theta(\vx))d\vx$, where $\theta$ denotes the weights that parameterize the energy function $E_\theta$. We will represent a potential as an EBM, defined using a neural network parameterized energy function $E_\theta(\vx): \mathbb{R}^D \rightarrow \mathbb{R}$ that maps each datapoint to a scalar value representing an energy. A potential then corresponds to the set of datapoints with low assigned energy. 
Thus, datapoints $\vx$ satisfying our petential have high likelihood, and all other datapoints have low likelihood. Potential satisfaction or minimization then corresponds to sampling from the EBM distribution $p_\theta(\vx)$.

\myparagraph{Minimizing Potentials.}

In our framework, minimizing a potential corresponds to sampling from the EBM which defines it, and thus finding high-likelihood data points under $p_\theta(\vx)$. We follow existing works and utilize a gradient based MCMC procedure, Langevin Dynamics \citep{Welling2011BayesianLV, Du2019ImplicitGA} to sample from the EBM distribution. In particular, to optimize a potential, we initialize a trajectory $\vx^0$ from uniform noise. We then run $M$ iterative steps following:
\begin{equation}
\resizebox{0.88\hsize}{!}{$
 \Tilde{\V{x}}^m =   \Tilde{\V{x}}^{m-1} - \frac{\lambda}{2}\nabla_{\V{x}}E_\theta \left(\Tilde{\V{x}}^{m-1} \right) + \omega^m, \quad \omega^m \sim \mathcal{N}(0, \sigma) 
 $}
 \label{eq:mixing},
\end{equation}
where at each step we iteratively optimize the trajectory with respect to the energy function, using an underlying gradient step size of $\lambda$ and noise scale of $\sigma$. We include hyperparameter details for sampling in Section \ref{sec:implementation} of the appendix, and heuristically set the noise scale of $\sigma=0$.

\subsection{Composing Potentials}
\label{sect:compose_constraint}

Next, we discuss how we may compose different sets of potentials together, where each potential is  parameterized by a separate EBM $E_\theta^j(\vx)$. Our composition operator builds on existing works on composing EBMs representing concepts \cite{Du2021UnsupervisedLO}.

\myparagraph{Sampling Composed Potentials.} Given a set of separate potentials, we wish to solve for a set of trajectories $\vx$ which jointly satisfy each of the potentials. In our EBM formulation, this corresponds to finding a trajectory $\vx$ which is low energy under each of the specified energy functions $E_\theta^j(\vx)$.

Such a setting is equivalent to finding a trajectory $\vx$  which has high likelihood under each EBM probability distribution $p_\theta^j(\vx)$. This corresponds to sampling from the distribution defined by the product of the individual EBM distributions,
\begin{equation}
\prod_j{p^j_\theta(\V{x})} \propto e^{- \sum_j {E_\theta^j{(\V{x})}}} = e^{- E_\theta'{(\V{x})}},
\end{equation}
which corresponds to a new EBM with energy function $E_\theta'{(\V{x})}$ (an analogous approach can be applied to generate images subject to a set of concepts \citep{Du2020CompositionalVG}). Thus, we may sample from the composition of a set of potentials using a sampling procedure as \eqn{eq:mixing}, using the new energy function $E_\theta'{(\V{x})}$, defined as the sum of each individual energy function. Intuitively, this corresponds to a continuous optimization procedure on each energy function.

In our setting, different energy functions $E_\theta^j(\vx)$ are constructed by conditioning an energy function on separate latent vectors. These latents are directly inferred without supervision from input trajectories by training an encoder jointly with the energy function parameters.

\section{Neural Interaction Inference with Potentials}
\vspace{-3pt}

\begin{figure*}[t!]
\centering
  \includegraphics[trim={100 590 210 0}, width=.9\textwidth]{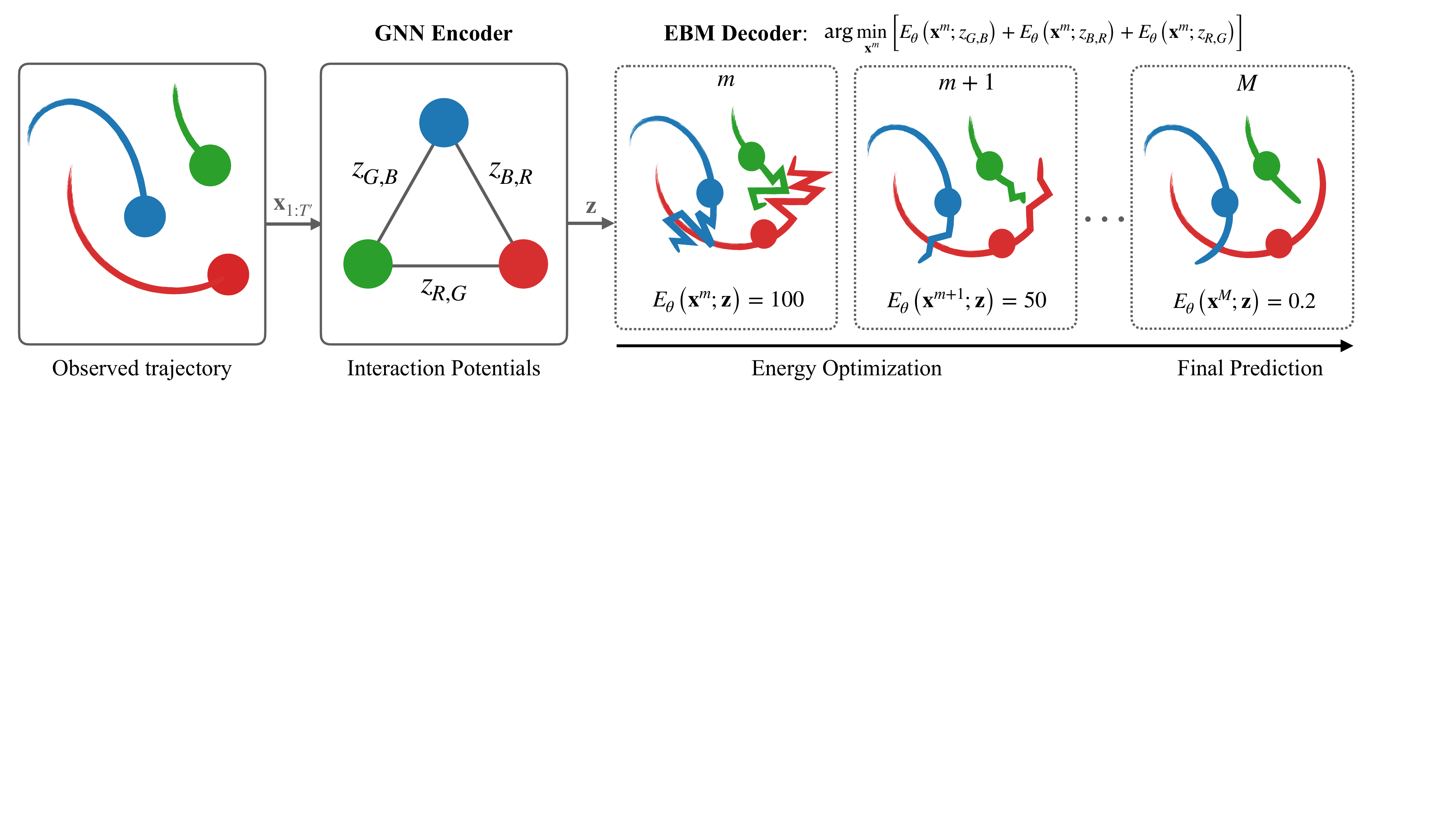}
  \caption{\small \textbf{Overview of \model.} In the left, a portion $\V{x}\left(1 \dots T^\prime \right)$ of the input trajectory is observed by $\text{Enc}_{\theta}$ and encoded by a GNN into interaction potentials, in the form of a set of latent vectors $\V{z}$ for each edge in the graph. In the right, energy functions parametrized as GNNs for each edge latent vector in $\V{z}$ are constructed. Energy functions are trained so that optimizing a trajectory $\V{x}^0$ from uniform noise into a final trajecotry $\V{x}^M$ reconstructs the future states of the observed trajectory. This refinement process uses Langevin Dynamics (Eq. \ref{eq:langevin-update}). Given the full trajectory $\V{x}^m$ at sampling step $m$, we update it by summing the gradient contributions of the energy function associated to each edge, resulting in $\V{x}^{m+1}$. } 
\label{fig:method}
\vspace{-14pt}
\end{figure*}

Next, we discuss Neural Interaction Inference with Potentials (\model), our  unsupervised approach to decompose a trajectory $\vx(1...T)_i$, consisting of $N$ separate nodes at each timestep,  into a set of separate EBM $E_\theta^j(\vx)$ potentials.  \model is composed by two steps: \textbf{(i)} an encoder for obtaining a set of potentials and \textbf{(ii)} a sampling process which optimizes for a predicted trajectory, minimizing the inferred potentials. Energy functions in \model are trained using autoencoding, similar to \cite{Du2021UnsupervisedLO}. We provide an illustration of our approach in \fig{fig:method}, pseudocode in Algorithm \ref{algo} and an illustration of the architecture in \fig{fig:method-app} of the Appendix.

\subsection{Relational Potentials}
\label{sect:energy_constraint}

To effectively parameterize different potentials for separate interactions, we learn a latent conditioned energy function $E_\theta(\vx, \vz) : \real^{T \times D} \times \real^{D_z} \xrightarrow{} \real$. Then, inferring a set of different potentials corresponds to inferring a latent $\vz \in \mathbb{R}^{D_z}$ that conditions an energy function.

Given a trajectory $\vx(1...T)_i$, we infer a set of $L$ different latent vectors for each each directed pair of interacting nodes in a trajectory. Thus, given a set of $N$ different nodes, this corresponds to a set of $N(N-1)L$ energy functions. 

\begin{figure}[h]
\begin{center}
 \scalebox{0.8}{
\begin{minipage}{.6\textwidth}
\begin{algorithm}[H]
\small
\begin{algorithmic}
    \STATE \textbf{Input:} Full trajectories $\V{x}$, Observed trajectories $\V{x}(1...T\prime)$, Initial conditions $\V{x}(1...T_0)$, step size $\lambda$, number of gradient steps $M$, Encoder $\text{Enc}_\theta$, energy functions $E_\theta^{ij,l}$, noise  $\omega^m = 0$, true data distribution $p_D$
    \WHILE{not converged}
    \STATE $\V{x}_i \sim p_D$
    \STATE \emph{$\triangleright$ Encode components $\V{z}_{ij,l}$ from $\V{x}(1...T^\prime)$}
    \STATE $\{\V{z}\} \gets \text{Enc}_\theta(\V{x}(1...T^\prime))$
    \STATE \emph{$\triangleright$ Optimize sample  $\V{x}_i^0$ via gradient descent:}
    \STATE $\V{x}^0_i \sim \mathcal{U}(0, 1)$ 
    \FOR{gradient step $n = 1$ to $N$}
    \STATE $\Tilde{\V{x}}^m \gets   \Tilde{\V{x}}^{m-1}- \frac{\lambda}{2}\nabla_{\V{x}}\sum_{ij,l}{E_\theta^{ij,l}(\Tilde{\V{x}}^{m-1};\V{z}_{ij,l})} + \omega^m$
    \ENDFOR
    \STATE \emph{$\triangleright$ Optimize objective $\mathcal{L}_{\text{MSE}}$ wrt $\theta$:} 
    \STATE $\Delta \theta \gets \nabla_\theta \|\Tilde{\V{x}}^m - \V{x}\|^2 $
    \STATE Update $\theta$ based on $\Delta \theta$ using optimizer 
    \ENDWHILE
  \end{algorithmic}
 \caption{Training algorithm for \model.}\label{algo}
 \end{algorithm}
 \end{minipage}}
 \end{center}
 \vspace{-5pt}
 \caption{\textbf{Training Algorithm.} \model is trained to infer a set of potentials, represented as energy functions, using a trajectory reconstruction objective. A set of latents $\{\vz\}$ is inferred from the beginning of a trajectory $\V{x}(1...T^\prime)$, and define different potentials. A trajectory is optimized w.r.t. to energy functions and supervised with the trajectory $\V{x}$. }
  \vspace{-15pt}
\end{figure}

To generate a trajectory, we optimize the energy function $E(\V{x}) = \sum_{ij,l} E_\theta^{ij,l}(\V{x};\V{z}_{ij,l})$, across node indices $i$ and $j$ from $1$ to $N$ and latent vectors $l$ from $1$ to $L$. However, assigning one energy function to each latent code becomes prohibitively expensive as the number of nodes in a trajectory increases. Thus, to reduce this computational burden, we parameterize $L$ energy functions as shared message passing graph networks, grouping all edge contributions $ij$ in a single network. The energy is then computed as a summation over all individual node energies after message passing. 
To evaluate the energy corresponding to a single edge factor $\V{z}_{ij,l}$ we mask out the contributions of all other edges to the final node energies. Architecture and further details can be found in Section \ref{sec:arch} of the appendix.

To condition to message passing shared graph network on each inferred latent $\vz_{ij,l}$, each edge $e(i,j)$ in the graph is conditioned by the corresponding encoded edge latent code $\V{z}_{ij,l}$, by means of FiLM modulation \citep{Perez2018FiLMVR}.

\subsection{Inferring Energy Potentials}
\label{sect:inference}

We utilize $\text{Enc}_{\theta}(\V{x}): \real^{T \times D} \xrightarrow{} \real^{D_z} $ to encode the observed trajectories $\V{x}$ into $L$ latent representations per edge in the observation. We utilize a fully connected GNN with message-passing to infer latents using the encoder module in \cite{Kipf2018NeuralRI}. Instead of classifying edge types and using them as a gate ouputs,  we utilize a continuous latent code $\V{z}_{ij,l}$, allowing for higher flexibility.

\subsection{Training Objective} 
\label{sect:nci}

To train \model, we infer a set of different EBM potentials by auto-encoding the underlying trajectory. In particular, given a  trajectory $\V{x}(1...T)_i=\left(\V{x}(1)_i,\dots,\V{x}(T)_i\right)$, we split the trajectory into initial conditions  $\V{x}(1...T_0)$, corresponding to the first $T_0$ states of the trajectory and $\V{x}(T_0...T)$, corresponding to the subsequent states of the trajectory, where each state of the trajectory consists of $N$ different nodes. The edge potentials are encoded by observing a portion of the overall trajectory $\V{x}(1...T^\prime)$, where $T^\prime \leq T$.

We infer a set of different $L$ latents per edge of input observations utilizing the observed states $\V{x}(1...T^\prime)$ using the encoder specified in \sect{sect:inference}, generating a set of latents $\{\V{z}\}$. We then aim to train energy functions so that the following unnormalized distribution assigns low energy and high likelihood to the full trajectory $\V{x}$:
\begin{align}
    p(\V{x}|\{\V{z}\}) &\propto \prod_{i,j,l \forall i \neq j} {p(\V{x}|\V{z}_{ij,l})} = 
    \nonumber \\ &= \text{exp}\left({-E_\theta^{ij,l}(\V{x};\text{Enc}_{\theta}(\V{x}(1...T^\prime))_{ij,l}}) \right),
    \label{eq:poe}
\end{align}
where $\V{z}_{ij,l} = \text{Enc}_{\theta}(\V{x}(1...T^\prime))_{ij,l}$ and $E_\theta^{ij,l}$ is the energy function linked to the $l_\text{th}$ potential of the encoded edge between nodes $i$ and $j$, respectively.

Since we wish to learn a set of potentials with high likelihood for the observed trajectory $\V{x}$, as a tractable supervised manner to learn such a set of valid potentials, we directly supervise that sample using \eqn{eq:mixing} corresponds to the original trajectory $\V{x}$, similar to \cite{Du2021UnsupervisedLO}. In particular, we sample $M$ steps of Langevin sampling starting from $\Tilde{\V{x}}^0$, which is initialized from uniform noise and the initial conditions fixed as the ground-truth $\vx(1...T_0)$:

\begin{equation}
 \Tilde{\V{x}}^m =   \Tilde{\V{x}}^{m-1}- \frac{\lambda}{2}\nabla_{\V{x}}\sum_{ij,l}{E_\theta^{ij,l}(\Tilde{\V{x}}^{m-1};\V{z}_{ij,l}} ) + \omega^m, \label{eq:langevin-update}
\end{equation}

where $m$ is the $m_\text{th}$ step and $\lambda$ is the step size and $\omega^m \sim \mathcal{N}(0, \lambda)$. We then compute MSE objective with $\Tilde{\V{x}}^M$, which is the result of $M$ sampling iterations and the ground truth trajectory $\V{x}$: 

\begin{equation}
      \mathcal{L}_{\text{MSE}}(\theta) = \| \Tilde{\V{x}}^M- \V{x} \|^2. 
\end{equation}
We optimize both $\Tilde{\V{x}}$ and the parameters $\theta$ with automatic differentiation. The overall training algorithm is provided in Algorithm \ref{algo}.

\section{Experiments}\label{sec:exp}

In this section we firstly describe our datasets (Section \ref{sec:datasets}) and baselines (Section \ref{sec:baselines}). Following, in Section \ref{sec:disentanglement}, we discuss experiments on \textbf{(i.)} recombination of interaction types across datasets and \textbf{(ii.)} contribution of the potentials. Next, in Section \ref{sec:ood}, we describe out-of-distribution sample detection experiments. We show how to incorporate test-time potentials in Section \ref{sec:flex-gen}. Finally, we describe the quantitative results  for trajectory forecasting in Section \ref{sec:quantitative}. 
In the appendix we give implementation details (Section \ref{sec:implementation}), experimental details (Section \ref{sec:experiment_det}), additional examples (Section \ref{sec:add_results}) and provide a detailed ablation study (Section \ref{sec:ablation}).

\subsection{Datasets}\label{sec:datasets}

We test our model in three different domains. First, we carry on experiments in two simulated environments: \textbf{(i.)} Particles connected by springs, and \textbf{(ii.)} Particles with charges. Next, we test several properties of our model in \textbf{(iii.)} NBA SportVu motion dataset, which displays real motion from tracked basketball players along several NBA games. Finally, we test our performance in \textbf{(iv.)} JPL Horizons, a physics-based realistic dataset.

\myparagraph{Simulated data.} Following the experimental setting described in \cite{Kipf2018NeuralRI}, we generate states (position and velocity) of a dynamical system for {$N=5$} particles for 70 time-steps. Our model observes the first 49 states, fixes one state and predicts the following 20. We generate 50k training samples and 10k for validation and test splits. 

In this setting, the rules by which particles interact are known and simple. However, they can generate very complex behaviour.
\begin{itemize}[leftmargin=*]
    \myitem \textbf{Springs}: The particles move inside a box with elastic collisions. They are connected by a spring with probability 0.5, and interact according to Hooke's law.
    \myitem \textbf{Charged}: The particles move inside a box as in Springs. They are assigned a positive or negative charge $q_i\in\{\pm q\}$ with probability 0.5 and interact via Coulomb forces.
\end{itemize}

\myparagraph{NBA SportVU} SportVU is an automated ID and tracking service that collects data of NBA players and the ball ({$N=11$}) during games. The inherent complexity of human motion and interactions while playing a sport makes this dataset especially challenging for forecasting. The dataset is generated by splitting each of the labeled events into 65 steps trajectories of coordinates x,y. We compute the velocities to generate the states. The dataset is composed of 50k samples for training and 1k samples for validation and test. 


\myparagraph{JPL Horizons}
The JPL Horizons on-line ephemeris system provides access to solar system data. It characterizes the 3D location and velocity of solar system objects (targets) as a function of time, as seen from locations within the solar system (origins). We choose this dataset as a realistic take on physical interactions. Here, inter-particle forces are a product of gravity, and therefore mass. However, we do not provide information about the mass or any other object attribute. There are other factors of complexity. First, there are hidden nodes (smaller objects) that are not visible to the observer, introducing noise to the trajectories. Second, the origin from which we observe the trajectories varies along samples.
This dataset consists on the trajectories captured between 1800 and 2022, with one datapoint every 10 days. We define the nodes as $N=12$ targets of the solar system: 8 planets, 3 moons and the Sun. This data is captured from 13 origins: each one of the targets plus the solar system barycenter (SSB). We gather 1880 trajectories of 43 timesteps split as 1504/188/188 for train, validation and test.

\subsection{Baselines} \label{sec:baselines}
We consider a \textbf{Static} baseline, which copies the previous state vector, a multi-node \textbf{LSTM} which is trained to predict the state vector difference at every timestep. It concatenates input representations from all objects after passing them through an MLP. 

We also evaluate \textbf{NRI}, the architecture presented in \cite{Kipf2018NeuralRI} to infer the interaction graph (\textbf{learned}), and with a fully connected graph of a single edge type (\textbf{full graph}). We further add a GNN conditioned to the observed trajectories. 

For NBA SportsVU we will also evaluate on two social interaction-based methods: Social-LSTM (S-LSTM) \cite{Alahi2016SocialLH} and Directional-LSTM (TrajNet++) \cite{Kothari2020HumanTF}, PwD \cite{Janner2022PlanningWD}, a diffusion-based planning method, and \textbf{dNRI} \cite{Graber2020DynamicNR}, an extension of NRI that allows for dynamic switches of edge-types.

\begin{figure}[t]
\centering
  \includegraphics[ trim={30 320 165 0}, width=0.85\textwidth]{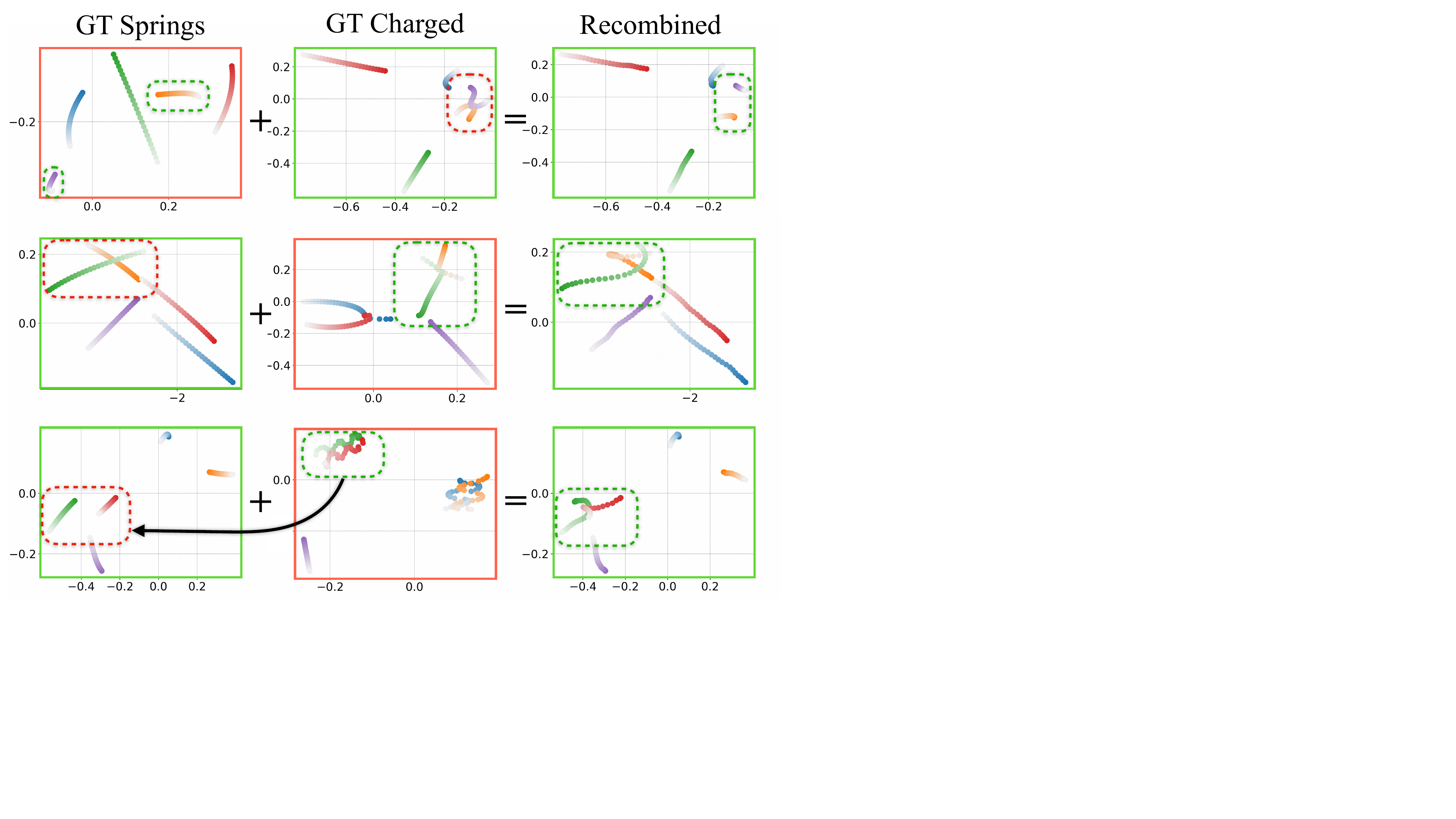}
  \caption{\small \textbf{\model can recombine encoded potentials at test-time learned from different datasets.} Illustrated, samples from Springs (Col. 1) and Charged (Col. 2) and their recombinations (Col. 3). \model encodes both trajectories. \model is able to reconstruct trajectories framed in green while swapping the edge potentials associated to the nodes in the green dashed box for the ones in the red dashed boxes. Recombinations look semantically plausible.}
\label{fig:recombination}
\vspace{-15pt}
\end{figure}
\subsection{Independence of Relational Potentials}\label{sec:disentanglement} 
\model assigns an energy function to each one of the representations learned by the encoder. Those constrain the generative process by conditioning the features of their associated EBM. The optimization procedure, hence, aims to minimize all potentials. We train our model to ensure that different potentials contribute independently to the generative process by addition of their associated gradients. Our training procedure enforces separation among \textbf{(i.)} Potentials associated to different relations and \textbf{(ii.)} Individual potentials affecting the same edges. 
We argue that our training procedure aids discovery of independent potentials allowing composition among disjoint training distributions.

\myparagraph{Recombination} 

To verify our claims, we show how \model can compose relational potentials learned from two different distributions, at test-time. Figure \ref{fig:recombination} shows qualitative results of recombinations from Springs and Charged datasets. 
The process is as follows: we train two instances of our model ($\text{\model}_S$, $\text{\model}_C$) to reconstruct Springs and Charged trajectories respectively. Given sample trajectories drawn from each dataset (Col. 1 for Springs and 2 for Charged), we encode them into their relational potentials. For each row, we aim to reconstruct the trajectory framed in green while swapping one of the potential pairs (green dashed box) with one drawn from the other dataset (red dashed box). As an example, in the first row of the figure, we encode the Springs trajectory with $\text{\model}_S$ and the Charged trajectory with $\text{\model}_C$. Next, we fix the initial conditions of the Charged trajectories and sample by optimizing the relational energy functions. To achieve recombination, each model targets specific edges. We minimize the potentials encoded by $\text{\model}_S$ for the mutual edges corresponding to the nodes in green dashed boxes. The rest of edge potentials are encoded by $\text{\model}_C$. The sampling process is done jointly by both models, each minimizing their corresponding potentials.
The result is a natural combination of the two datasets, which affect only the targeted edges. Reconstructed trajectories in Figure \ref{fig:recombination} (col. 3) are semantically reasonable.

\myparagraph{Contribution of Potentials}

\begin{figure}[t]
\centering
  \includegraphics[ trim={65 610 870 0}, width=.412\textwidth]{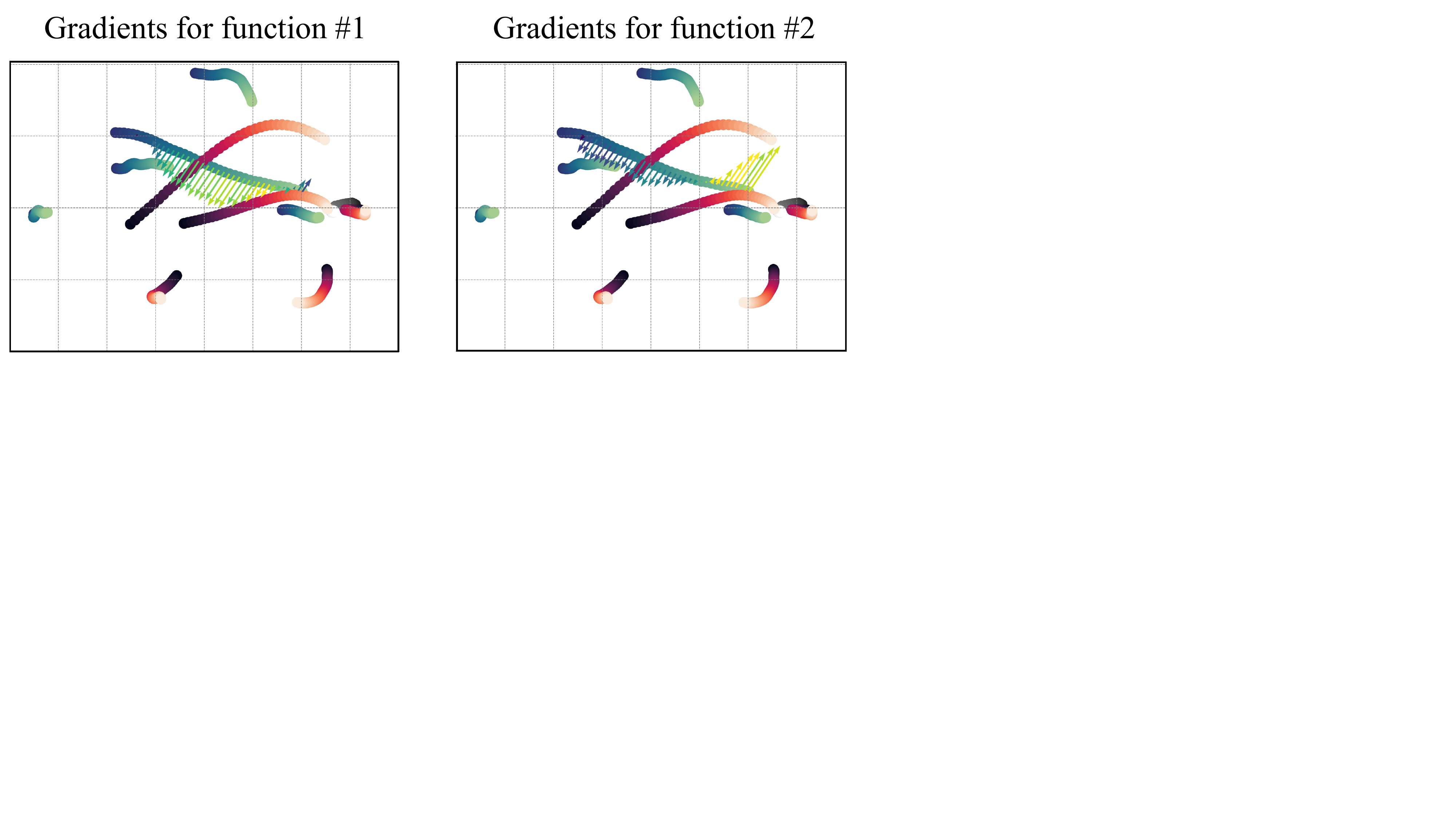}
  \vspace{-3pt}
  \caption{\small \textbf{Energy potentials discovered by our approach control different aspects of the trajectories}. For a model trained with 2 energy functions, this illustration shows the gradients associated to each energy function applied to a ground-truth sample. Each potential pushes the player of interest into a different direction. }\label{fig:gradients}
  \vspace{-10pt}
\end{figure}

As introduced, we can assign more than one potential to each edge. We argue that each one of those $L$ potentials will control different aspects of the same interaction. A qualitative example shown in Figure \ref{fig:gradients} depicts the gradient orientation of two sets of potentials evaluated in a single node. We can see how each potential pushes the player trajectory into different directions, each one of them pointed to a different player of the rival team.

\begin{figure}
    \centering
\includegraphics[ trim={20 620 700 0}, width=0.5\textwidth]{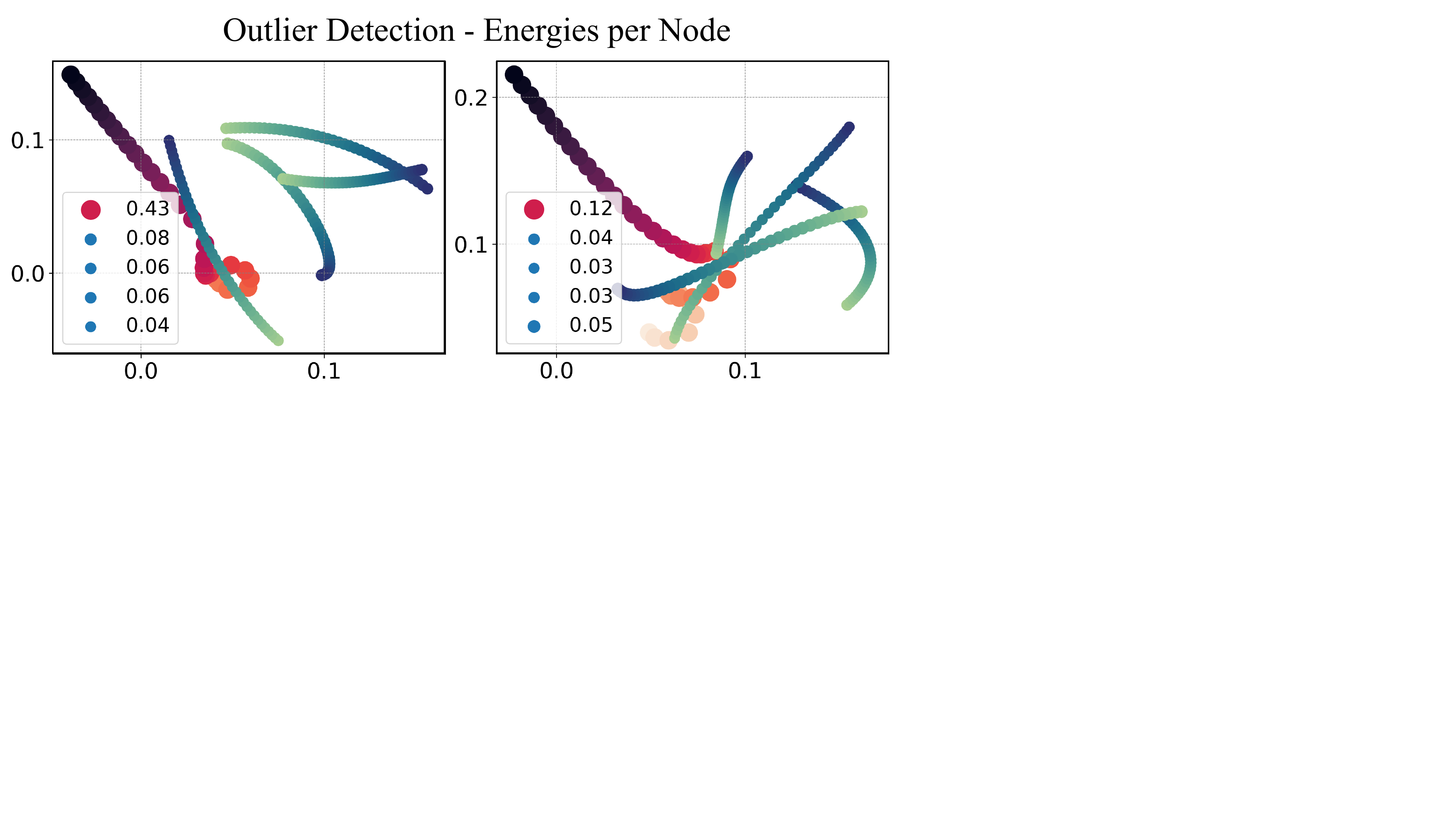}
    \caption{\small \textbf{Out-Of-Distribution Detection with \model}. A model trained with certain relation types can detect when a trajectory exhibits a new type of relation. The illustrated trajectories show the energy associated to each one of the nodes. Red trajectories: Charged particles, Blue-Green trajectories: Springs particles. We train \model in Springs dataset. Our model assigns higher energies to those nodes that behave differently than the training set.}\label{fig:ood}
    \label{fig:my_label}
\vspace{5mm}
\hspace{-5mm}
\begin{minipage}[t]{0.46\linewidth}
\scalebox{0.64}{
\begin{tabular}{l|c}
    \toprule
    \textbf{Evaluation} &  \textbf{Train: Springs (S)}\\ \midrule
    & \textbf{Node energy} \\ \midrule
    \textbf{Springs (S)}     & 5.1e-3 \\ %
    Charged (C)           & 1.8e-1  \\ %
    S\&C   (eval S)    & 9.1e-2  \\%
    S\&C   (eval all) & 1.4e-1  \\%
    S\&C   (eval C)    & 1.9e-1 %
    \vspace{2mm}
\end{tabular}}

\end{minipage}
\hspace{-1mm}
\begin{minipage}[t]{0.49\linewidth}
\centering
\scalebox{0.62}{
\begin{tabular}{l|c}
    \toprule
      \textbf{Evaluation} &   \textbf{Train: NBA Players (P)} \\ \midrule
        & \textbf{Node energy} \\ \midrule
    \textbf{P\&Ball  (eval P)}         & 5.9e-2  \\%
    P\&Ball (eval all)       & 8.4e-2  \\%
    P\&Ball (eval Ball)       & 3.2e-1  \\%
    \midrule 
       & \textbf{Detection Accuracy}  \\ \midrule 
        Ball      & 70.1$\%$ 
\end{tabular}}
\end{minipage}
\captionof{table}{\textbf{Quantitative evaluation of out-of-distribution detection}.  We evaluate the average energy associated to each node-type in a scene. In the left, \model is trained on Springs and evaluated onn (i) Springs (ii) Charged and (iii) S\&C, a Springs-Charged mixed dataset. For the NBA case in the right, we train \model for the subset of player (P) trajectories of the dataset and evaluate the energies in a setting with player nodes and one ball node. We measure accuracy in detecting the ball trajectory.}\label{tab:ood}

\vspace{-20pt}
\end{figure}

\subsection{Out-Of-Distribution Detection} \label{sec:ood}

We further utilize the potential value or energy produced by \model over a trajectory to detect out-of-distribution interactions in a trajectory. In our proposed architecture, energy is evaluated at the node level. Therefore, if \model has been trained with a specific dataset, the potentials associated to out-of-distribution type of edges are expected to correspond to higher energy. 

We design a new dataset (Charged-Springs) as a combination of Springs and Charged interaction types. In simulation, nodes are assigned both roles of Charged and Springs particles, but all the forces they receive correspond to one of the two types with probability $p=0.5$. We train a model with the Springs dataset and evaluate the energies in the proposed mixed setting.

Figure \ref{fig:ood} shows qualitatively how the energy is considerably higher for the nodes with Charged-type forces (drawn in red). Quantitative results are summarized in Table \ref{tab:ood} for 1k test samples. In the left, we can see that energies corresponding to Spring-type nodes are considerably lower than for Charged-type nodes, indicating that potentials are correctly capturing the behavior of the desired interactions. 

We further evaluate the OOD detection for NBA SportsVU dataset. For this experiment, we train \model with the 10 players disregarding the Ball node. At test-time, we evaluate the trained model switching one of the players by the Ball node for 1k samples. We observe how the energy corresponding to the Ball is considerably higher. We further train a single parameter binary classifier and find that we can detect the Ball in 70.1\% of instances (Table 1 right).

\begin{table}[t]
\small
\centering
  \includegraphics[ trim={10 660 150 0}, width=.9\textwidth]{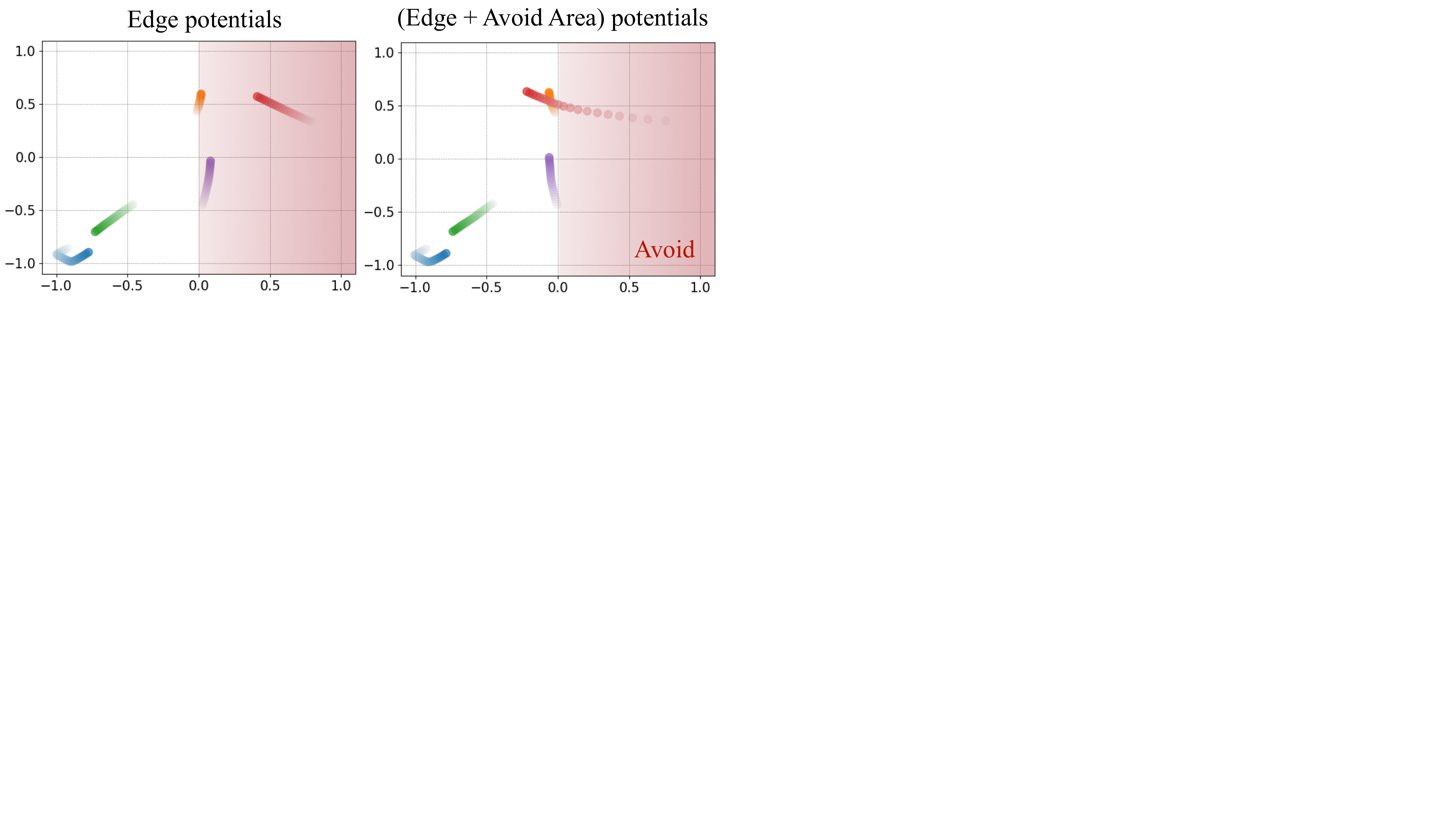}

\scalebox{0.66}{
\begin{tabular}{l|cccccc}
Potentials & GT & NRI & \model & +$P_{add}$ s1 & +$P_{add}$ s2 & +$P_{add}$ s3  \\ \midrule 
Goal: Squared Distance & 7.53e-1 & 7.52e-1 & 7.52e-1 & 7.02e-1 & 6.18e-1 & 3.89e-1 \\ 
Avoid Area: $\%$ in area & 49.6$\%$ & 49.6$\%$& 49.5$\%$ & 48.3$\%$ & 31.7$\%$ & 21.8$\%$
 \\
\bottomrule
\end{tabular}
}\captionof{figure}{\small \textbf{With \model we can add and control test-time potentials to achieve a desired behavior.} We design test-time potentials to steer trajectories into a goal. In row 1 of the table, we show the squared distance after applying a goal potential towards the center (0,0) of the scene with different strengths. In row 2, we report the percentage of time-steps that particles stay in a particular area $\V{A} =[(0,-1), (1,1)]$ after applying a potential that enforces avoiding $\V{A}$. The figure shows the effect of applying test-time potentials in the latter experiment.}\label{tab:charged_new_potentials}
\vspace{-15pt}
\end{table}

\subsection{Flexible Generation} \label{sec:flex-gen}

\begin{figure}[t]
\centering
  \includegraphics[ trim={20 80 440 0}, width=0.9\textwidth]{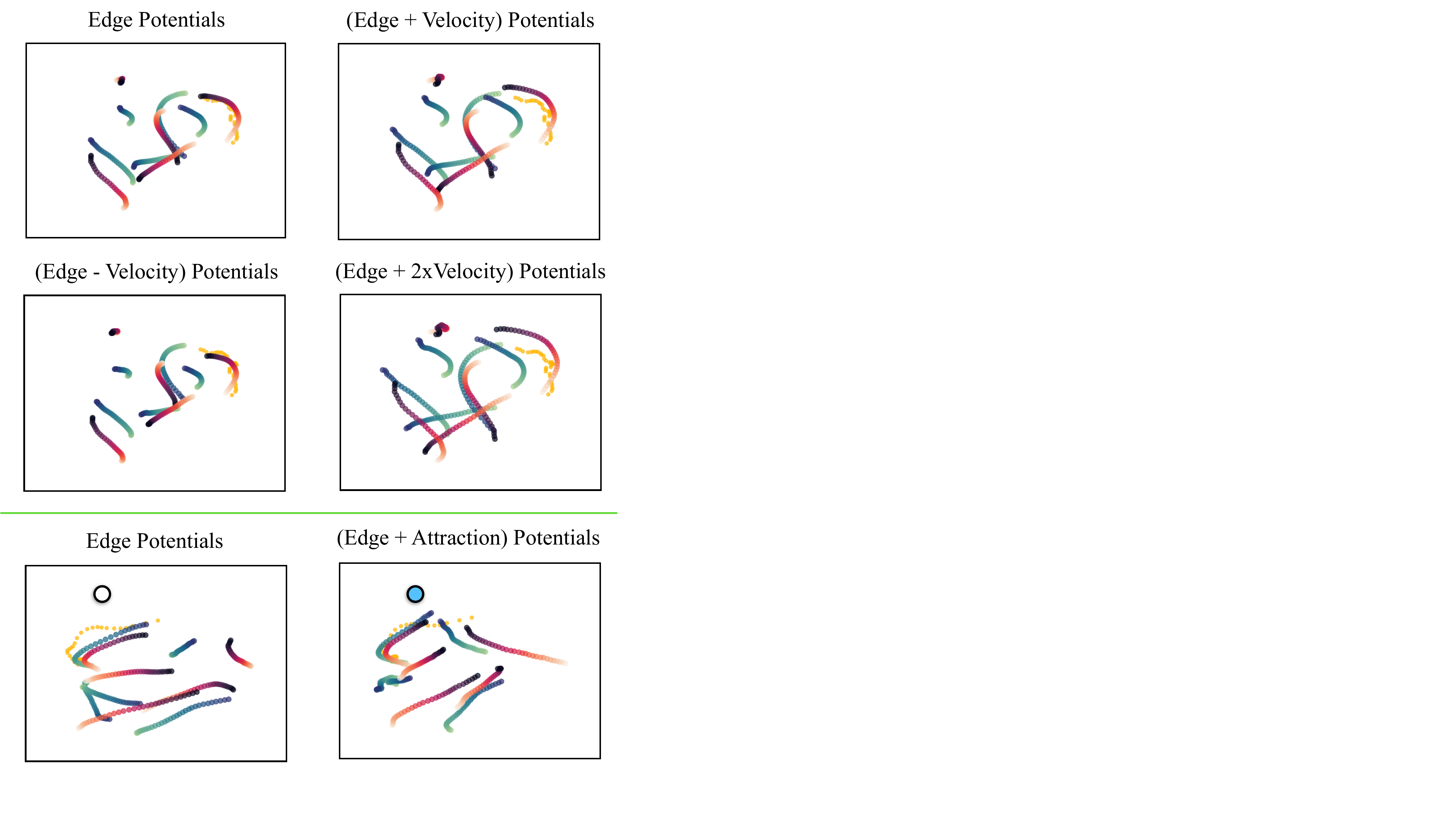}
  \caption{\small \textbf{\model is able to incorporate new potentials in test-time}. We can see depicted reconstructions of NBA samples with added potentials. Left: (Col. 1, Row 1): Reconstruction of the encoded trajectory. (Col. 1, Row 2): Decrease of velocity. (Col. 2, Row 1): Low increase of velocity. (Col. 2, Row 2): High increase of velocity. Right: (Col. 1, Row 3): Reconstruction of the encoded trajectory, (Col. 2, Row 3): Attraction of the players to a goal point (blue dot). Painted orange, the ground-truth ball trajectory.}
\label{fig:new-constraints}
\vspace{-15pt}
\end{figure}

Another advantage of our approach is that it can flexibly incorporate test-time user specified potentials. For this experiment, we investigate three different sets of potentials. We do so qualitatively by reconstructing a given 40 step trajectory of the NBA dataset in Figure \ref{fig:new-constraints}, and also quantitatively in Table \ref{tab:charged_new_potentials} for 20 step trajectory prediction.

\myparagraph{Velocity Potentials} We incorporate the following velocity potential as an energy function: $E = \epsilon \lambda \sum_{i,t}\sqrt{(\V{v}_{x,i}^t)^2 + (\V{v}_{y,i}^t)^2} = \epsilon \lambda\sum_{i,t}mod(\V{v}_{i}^t)$, for particle $i$ in time $t$. The weight $\lambda =1e-2/{N}$ scales the effect of this function over the rest and $\epsilon$ is a multiplicative constant that indicates the strength and direction of the potential. Figure \ref{fig:new-constraints} (two top rows), we show (\textbf{i.}) $\epsilon = 0$: Reconstruction (top-left); (\textbf{ii.}) $\epsilon = 4$: Decrease of velocity (middle-left); (\textbf{iii.}) $\epsilon = -5$: Low increase of velocity (top-right); and (\textbf{iv.}) $\epsilon = -10$: High increase of velocity (middle-right). Results satisfy test potentials.

\myparagraph{Goal Potentials} We also add at test-time an attraction potential as a the squared distance of the predicted coordinates to the goal: $P = \epsilon\lambda \sum_{i,t}(\Tilde{\V{p}}^{0:T-1}_i - \V{g})^2$, where $\V{g}$ is defined as the coordinates of our goal point. We define the trajectory coordinates $\Tilde{\V{p}}^{t}_i$ as an accumulation of the un-normalized velocities $\V{v}^{t}_i$ predicted by \model: $\Tilde{\V{p}}^{t+1}_i = \sum_{t}(\V{v}^{0:t}_i) + \V{p}^0_i$ for particle $i$ at time-step $t$.  Here, $\V{p}^1$ is fixed initial ground-truth location of the particle at time 0 and $\lambda =5e-4/{N}$.

Figure \ref{fig:new-constraints} (bottom row) illustrates the scenarios (\textbf{i.}) Reconstruction (bottom-left) and Attraction to the goal (bottom-right, goal in blue). The reconstructed trajectory follows the new potential, while maintaining the potentials of the encoded trajectory. 

In Table \ref{tab:charged_new_potentials} (row 1), we explore quantitatively the effect of different magnitudes of the added goal potential for prediction in the Charged dataset. Our test set is composed by 1k samples.
For this experiment, the particles live within a $[-1,1]$ box, for both $x$ and $y$ coordinates. The goal is the center $\V{g} = (0,0)$. In the table, $P_{add}$ indicates the use of edge potentials encoded by \model while adding the new potentials with different strengths: (\textbf{i.}) $s1: \epsilon=1$, (\textbf{ii.}) $s2: \epsilon=5$ and (\textbf{iii.}) $s3: \epsilon=10$. We observe how the squared distance to the goal decreases as expected.

\vspace{-5pt}
\myparagraph{Avoid Area Potentials} In this case, we penalize the portion of the predicted trajectory $\Tilde{\V{p}}^{t}_i$ that is inside a given restricted area $\V{A}$. We do so by computing the distance of each particle $i$ that lays within the region $\V{A}$ to the borders $\V{b}_A$ of $\V{A}$. The added potential is: $P = \epsilon\lambda \sum_{i,t}(\Tilde{\V{p}}_{A,i}^{0:T-1} - \V{b}_A - C)^2$, where $C$ is a small margin that ensures that the particles are repelled outside of the boundaries of $\V{A}$. With $\lambda =1e-3/{N}$.

In Table \ref{tab:charged_new_potentials} (row 2), we explore the effect of different strengths of this potential type, in the prediction task. We also provide a visual example. For this experiment, we avoid the area $\V{A} = [(0, -1), (1, 1)]$, which corresponds to half of the box (see Figure). We use the following parameters: (\textbf{i.}) $s1: \epsilon=1$, (\textbf{ii.}) $s2: \epsilon=5e1$ and (\textbf{iii.}) $s3: \epsilon=5e2$.

\subsection{Quantitative Comparison} \label{sec:quantitative}

In this Section, we aim to assess \model's capability for trajectory forecasting.
For all datasets, we will observe a portion of the trajectory and predict $20$ timesteps. 

We first test our approach in Springs and Charged datasets. We evaluate the Mean-Squared Error (MSE) against \cite{Kipf2018NeuralRI}, their chosen baselines, a generic Conditional-GNN. Our models observes $49$ timesteps and fixes the $50_{th}$ as initial conditions for prediction. We can see in Table \ref{tab:pred_mse} that \model achieves better prediction error in both datasets. 

Similarly, for NBA (Table \ref{tab:pred_mse_nba_jpl} (left)) the model observes $40$ timesteps and fixes the following $5$ as initial conditions for prediction. \model outperforms the baselines in terms mid and long-term prediction error, and underperform in the short-term. 

The models designed for social interaction perform poorly in long-term prediction, while they have shown to excel in other tasks such as collision avoidance. 

\begin{table}[t!]
\small
\centering
\scalebox{0.76}{
\begin{tabular}{l|ccc|ccc}
\toprule
& \multicolumn{3}{c}{\textbf{Springs}} & \multicolumn{3}{c}{\textbf{Charged}} \\ \midrule
Prediction steps & 1 & 10 & 20 & 1 & 10 & 20 \\ \midrule
Static   & 1.70e-3 & 2.71e-2 & 2.55e-2 & 5.09e-3 & 2.30e-2 & 5.55e-2 \\

LSTM       & \textbf{1.10e-7} & 2.07e-6 & 4.65e-5 & 
            \textbf{9.69e-4} & 5.43e-3 & 1.15e-2\\
{Cond. GNN} & {1.35e-5} & {2.21e-5} & {3.44e-5} & 
                {3.67e-3} & {5.61e-3} & {1.05e-2} \\
NRI (full graph)     & 6.10e-4 & 7.82e-3 & 1.01e-2 & 
                    1.59e-3 & 3.52e-3 & 7.74e-3 \\
NRI (learned)     & 5.81e-7 & 1.10e-5 & 2.90e-5 &  
                1.47e-3 & \textbf{3.19e-3} & 6.65e-3\\
\model (Ours)   & 1.99e-7 &  \textbf{1.20e-6} &  \textbf{2.71e-6} & 
                \textbf{9.38e-4} & \textbf{3.07e-3} & \textbf{5.97e-3} \\ %
\bottomrule
\end{tabular}
}
\vspace{-5pt}
\caption{\small \textbf{Mean squared error (MSE) in predicting future states for Springs and Charged simulation datasets}, with 5 interacting objects. \model outperforms existing methods.}
\label{tab:pred_mse}
\vspace{-15pt}
\end{table}

FFor the JPL Horizons dataset in Table \ref{tab:pred_mse_nba_jpl} (right), \model outperforms the baselines also in mid and long-term prediction. Models have access to 23 timesteps. \model observes $20$ timesteps and fixes $3$ as initial conditions for prediction. JPL Horizons is a challenging dataset given the unknown masses of the bodies involved, as well as effects of unobserved smaller bodies nearby them. 
\section{Literature} \label{sec:rel-work}

\paragraph{Dynamics and Relational Inference} Several works in the past years have studied the problem of learning dynamics of a physical system from simulated trajectories with graph neural networks (GNNs) \citep{ Guttenberg2016PermutationequivariantNN, Gilmer2017NeuralMP, Steenkiste2018RelationalNE, lu2021learning, li2018learning, yang2022generative, rubanova2021constraint}. 

As an extension of the foundational work of \cite{Battaglia2016InteractionNF}, interaction networks, \cite{Kipf2018NeuralRI} proposes to infer an explicit interaction structure while simultaneously learning the dynamical model of the interacting systems in an unsupervised manner, by inferring edge classes with a classifier. Selecting models based on observed trajectories is also the base of \cite{Alet2019NeuralRI,Goyal2019RecurrentIM,Graber2020DynamicNR, fnri}. \cite{Graber2020DynamicNR} extends \cite{Kipf2018NeuralRI} to temporally dynamic edge constraints, which yields better results in real-world datasets. \model differs from these approaches as the generation procedure uses an optimization solver to satisfy a set of potentials, while learning relation representations as potentials from observation. \model models trajectories in the absence of attributes, by observing particles behave and without supervision.
Similarly, in \cite{Goyal2021NeuralPS} interactions are encoded as condition-action rules, which offer dynamics decomposition and modularity but lack the illustrated properties that energy functions offer.

\myparagraph{Energy-Based Models} Energy-based models have a long history in machine learning. Early work focuses on density modeling \cite{Hinton2002TrainingPO, Du2019ImplicitGA} by aiming to learn a function that assigns low energy values to data that belongs to the input distribution. 
To successfully sample data-points, EBMs have recently relied gradient-based Langevin dynamics \cite{Du2019ImplicitGA}. Recent works have illustrated that such a gradient-based optimization procedure can enable the composition of different energy functions \cite{Du2020CompositionalVG} and can successfully be applied to high-dimensional domains such as images \cite{Liu2021LearningTC, zhang2022robust, Nie2021ControllableAC}, trajectories \cite{urain2021composable, Du2019ModelBP}, and concepts \cite{Wu2022ZeroCAN}. 
In \cite{rubanova2021constraint} an energy approach is used to model trajectories. They make use of known object attributes and a global energy landscape to generate trajectories, while they do not discover interaction representations. \model leverages the properties of energy functions to learn energy potentials that can be composed in different ways.
Unsupervised discovery of composable energy functions has been previously explored on images \cite{Du2021UnsupervisedLO, zhang2022robust}. In this work, we extend ideas of unsupervised concept learning in EBMs to potentials and apply them to dynamical modelling and relational inference.

\begin{table}[t!]
\small
\centering
\scalebox{0.75}{
\begin{tabular}{l|ccc|ccc}
\toprule
& \multicolumn{3}{c}{\textbf{NBA SportsVU}} & \multicolumn{3}{c}{\textbf{JPL Horizons}} \\ \midrule
Prediction steps & 1 & 10 & 20 & 1 & 10 & 20 \\ \midrule
S-LSTM & 6.60e-5 & 6.67e-3 & 2.57e-2 & - & -& -\\
TrajNet++ & 5.30e-5 & 5.88e-3 & 2.33e-2 & -& - & - \\
Static & 2.13e-4 & 3.04e-3 & 1.07e-2    & 3.33e-3 & 5.54e-2 & 9.05e-2 \\ %
LSTM      & 8.07e-5 & 1.42e-3 & 5.31e-3 & 1.97e-6 & 3.98e-5 & 1.09e-4\\ %
PwD & 2.58e-4 &  1.17e-3 & 3.33e-3 & 8.38e-3 & 8.72e-3 & 9.04e-3 \\				
{Cond. GNN} & 1.71e-4 & 1.12e-3 & 3.11e-3 & 4.57e-6 & 4.66e-6 & 5.96e-6 \\
{NRI (learned)} &  {3.56e-6} & {7.46e-4} & {2.74e-3} & {2.67e-7} & {7.35e-7} & {1.16e-6} \\
{dNRI } & {\textbf{2.11e-6}} & {9.11e-4} & {3.52e-3} & \textbf{1.51e-7} &  {4.51e-6} & {2.26e-5} \\
\model (Ours)  & 3.15e-5 & \textbf{5.84e-4} & \textbf{2.37e-3} & 2.98e-7 & \textbf{4.32e-7} & \textbf{5.84e-7} \\ 
\bottomrule
\end{tabular}
}
\vspace{-5pt}
\caption{\small \textbf{Mean squared error (MSE) in predicting future states for NBA dataset and JPL Horizons dataset}, with 11 and 12 interacting objects respectively. \model performs better than the baselines at mid to long terms.}\label{tab:pred_mse_nba_jpl}
\vspace{-25pt}
\end{table}

\section{Discussion}
\vspace{5pt}

\myparagraph{Limitations.} Our existing formulation of \model has several existing limitations. First, \model is currently limited to encoding energy potentials associated with edges in graph neural networks. 

In practice, many potentials in nature are often not just simply pairwise, but depend on multiple sets of different particles. 

An interesting direction of future work is to explore how to generalize the use of the energy functions to capture multi-node interactions in a graph.

In addition, we found that relational potential discovered by \model were not necessarily cleanly disentangled. When recombining potentials between two datapoints with large differences, we sometimes found that interactions would be incorrectly generated. Similarily, we found that discovered potentials would erroneously assign high energy to particle interactions exhibiting the correct interaction. We believe that some of these issues may be rooted in the potential functions capturing unwanted information along with relational information (such as trajectory shapes). 
As a result, for instance, swapping relation potentials may produce unrealistic interactions. In the future, we believe that explicitly enforcing our encoding function to discard all trajectory information other than the relationship types may lead to improved performance.

\myparagraph{Conclusion.} In this work we introduced Neural Interaction Inference with Potentials (\model) which infers relational potentials specified as energy functions to model the dynamics of an interacting system. We illustrate how \model 
 as an alternative approach to discover such interactions that enables greater flexibility in trajectory modeling: it discovers a set of relational potentials, represented as energy functions, which when minimized reconstruct the original trajectory.
Throughout this work we explore and test the different advantages that our approach brings to trajectory modeling.
Particularly, we show that \model displays unique capabilities in test-time. It allows trajectory manipulation, such as interchanging interaction types across separately trained models. \model can also detect out-of-distribution samples without having been trained to do so, by observing the energies that correspond to each particle. We further show how we can modify the behavior of the modeled trajectories by adding test-time potentials. Finally, \model can also predict trajectories faithfully in the future, displaying favorable mid- and long-term performance when compared to existing approaches.

\section{Acknowledgements}
This  work was supported by NSF grants IIS-1814631 and CNS--2038493, AFOSR grant FA9550-19-1-0005, and ONR grant N00014-21-1-2431. Yilun Du is supported by a NSF Graduate Fellowship.

\bibliography{icml2023_conference}
\bibliographystyle{icml2023}

\newpage
\appendix
\onecolumn
\section{Appendix}

\renewcommand{\thefigure}{A\arabic{figure}}
\renewcommand{\thetable}{A\arabic{table}}

\setcounter{figure}{0}
\setcounter{table}{0}

In this appendix, we present implementation details in Section \ref{sec:implementation}. Details about the architecture and training procedure can be found in Section \ref{sec:arch}. Next, experiment settings are described in Section \ref{sec:experiment_det}. Following, a complexity analysis can be found in Section \ref{sec:comp-analysis}. Additional qualitative results with generated trajectories can be found in Section \ref{sec:add_results}, together with additional quantitative results for downstream tasks and new prediction experiments.
We further run an ablation study of each of our proposed components in Section \ref{sec:ablation}, together with an ablation study in test-time. 

Finally we discuss the broader impact of our work in Section \ref{sec:broad}.

\subsection{Implementation details} \label{sec:implementation}

\paragraph{Software:} We implemented this method using Ubuntu 18.04, Python 3.6, Pytorch 1.10, Cuda 11.2 and several additional libraries which will be provided as a environment requirements file.
\paragraph{Hardware:} For each of our experiments we used 1 GPU RTX 2080 Ti (Blower Edition) with 12.8GB of memory. Models are trained for approximately 1 day.

\subsection{Architecture and Training Details}\label{sec:arch}
In this section we discuss in depth the architecture of the main modules of our method. We also discuss the idiosyncrasies of our training procedure.

\paragraph{Architecture}
The architecture of \model is composed of 3 main modules: \textbf{(i.)} The encoder in Figure \ref{fig:encoder} is composed by convolutional and multi-layer perceptron blocks, with ELU activation functions. It encodes the observable trajectory $\V{x}(1...T^\prime)$ into a set of $L$ latent codes per edge, with a total of $N\times(N-1)$ edges. \textbf{(ii.)} The short-term energy function in Figure \ref{fig:trans-ebm} processes the trajectory in chunks of 5 time-steps. \textbf{(iii.)} The long-term energy function in Figure \ref{fig:conv-ebm} processes the trajectory with several convolutional filters, while reducing its temporal resolution. It finally temporally pools the whole trajectory. It is designed to observe the overall shape of the trajectory. Both energy functions make use of the Swish activation function. 
The resulting energy is the summation of the short and long-term energies $E = E_{LT} + E_{ST}$. The terms node $\rightarrow$ edge and edge $\rightarrow$ node correspond to the different steps of message passing procedure. In node $\rightarrow$ edge, information from a connected node pair is concatenated in an edge representation. edge $\rightarrow$ node represents the summated contribution of all edge features connected to every node. The conditioning blocks modulate the energy function features by means of FiLM modulation \cite{Perez2018FiLMVR}.

\begin{figure*}[t!]
\centering
  \includegraphics[trim={10 540 800 0}, width=.8\textwidth]{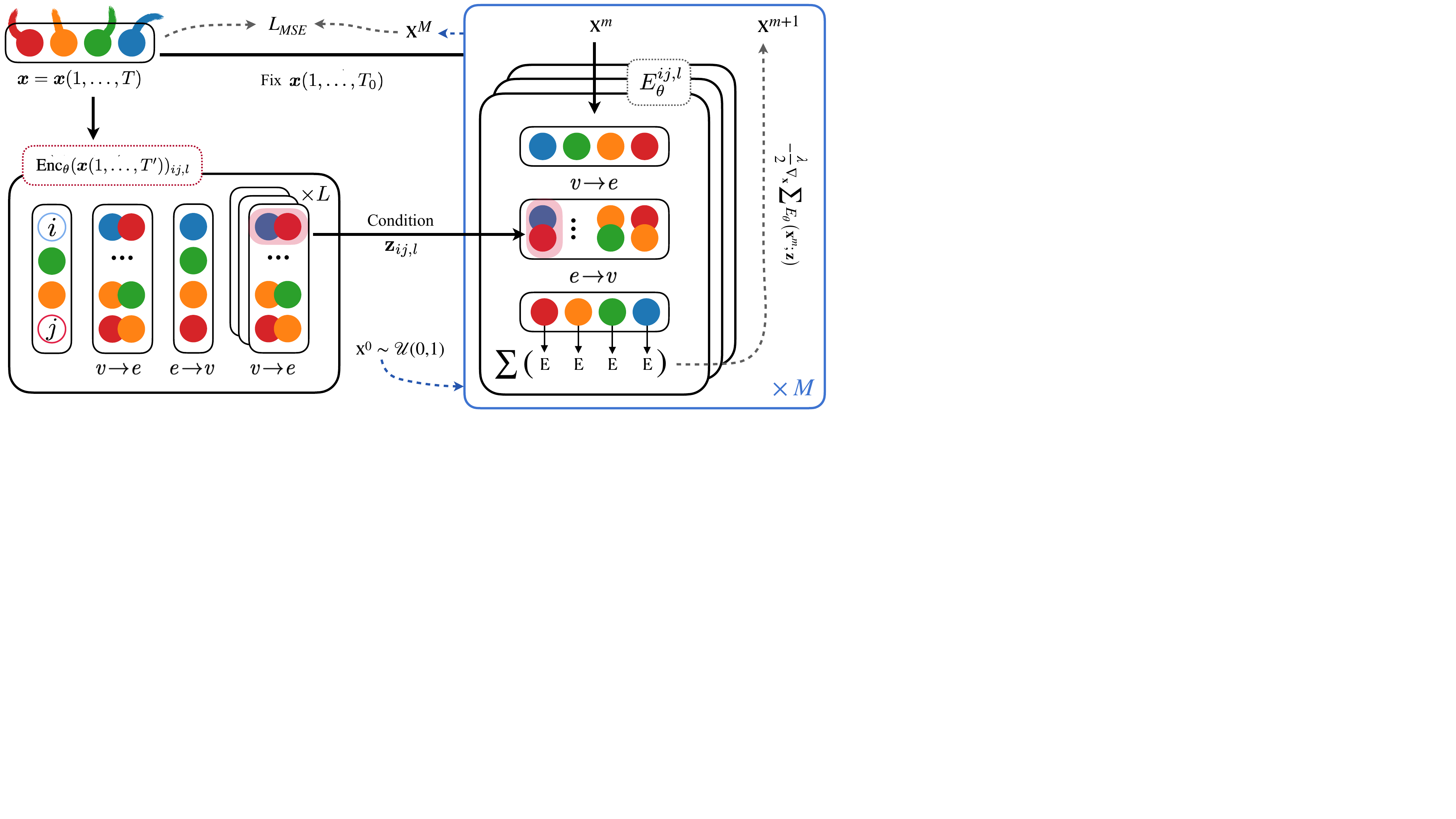}
  \caption{\small \textbf{Architecture and methodology of our approach.} In the left, $\text{Enc}_{\theta}$ observes a portion of the input trajectory $\V{x}$ and encodes them into potentials, in the form of latent vectors $\V{z}_{ij,l}$. In the right, a set of energy functions parametrized as GNNs are conditioned by $\V{z}_{ij,l}$ at the edge level. We initialize a trajectory $\V{x}^0$ as uniform noise and the ground-truth initial conditions $\V{x}(1..T_0)$, and update it by minimizing the energy functions associated to the encoded potentials. We sample by means of Langevin Dynamics. We supervise the reconstructed trajectories with an MSE objective with respect to the ground-truth trajectory.}
\label{fig:method-app}
\end{figure*}

An illustration of the overall architecture can be seen in Figure \ref{fig:method-app}.

\paragraph{Potential Splitting}

To generate a trajectory, we optimize the energy function $E(\V{x}) = \sum_{ij,l} E_\theta^{ij,l}(\V{x};\V{z}_{ij,l})$, across node indices $i$ and $j$ from $1$ to $N$ and latent vectors $l$ from $1$ to $L$. Explicitly computing one energy function per edge becomes prohibitively expensive as the number of nodes in a trajectory increases. As introduced, the computational burden is reduced by utilizing a shared message passing graph network to compute a fixed set of features for all edges (Tables \ref{fig:conv-ebm}, \ref{fig:trans-ebm}, \ref{fig:encoder} and Figure \ref{fig:method-app}). Hence, the energy corresponding to a single edge factor $\V{z}_{ij,l}$ is obtained by masking out the contributions of all other edges. 
However, in order to recombine edge types across multiple datasets it is desirable to train the model to combine multiple energy function contributions. With this objective, in training time we randomly split the encoded edge potentials into two disjoint subsets. The generated trajectory is a product of joint optimization of two energy functions, each one conditioned to one of the subsets. Each energy function observes one subset of potentials while masking out the contributions of the rest of edges.

\paragraph{Regularization}
To speed up training and regularize the energy values, we found useful to add the Contrastive Divergence loss $\mathcal{L}_{\text{CD}}$ \cite{Hinton2002TrainingPO}:
\begin{equation}
    \mathcal{L}_{\text{CD}} = \E_{p_D(\V{x})} \left[\sum_{ij,l}{E_\theta^{ij,l}(\V{x};\V{z}_{ij,l}} ) \right] - \E_{\text{stop\_grad}(q_{\theta}(\Tilde{\V{x}}))} \left[\sum_{ij,l}{E_\theta^{ij,l}(\Tilde{\V{x}};\V{z}_{ij,l}} )\right],
    \label{eq:cd_loss}
\end{equation}
{where $p_D(\V{x})$ is the true distribution of the data, and $q_{\theta}$ the distribution approximated by \model.}
We also regularize the energy values by penalizing the squared energy resulting from above. These regularizations are not necessary for the successful training of our model. However, they are helpful to stabilize training and therefore used for all experiments.
Both regularizations are added to the primary objective (MSE) with a weight of $\lambda_{reg}= 1e-4$.

\scalebox{0.81}{
\begin{minipage}{0.4\linewidth}
\centering
\begin{tabular}{c}
    \toprule
    \toprule
    Node $\rightarrow$ Edge \\
    \midrule
    5x1 CNN Block Down (2) 64\\ 
    \midrule
    CNN Conditioning Block (2) 64 \\
    \midrule
    CNN Conditioning Block (2) 64 \\
    \midrule
    Temporal Avg. Pool\\
    \midrule
    Edge $\rightarrow$ Node \\
    \midrule
    MLP 64 \\
    \midrule
    Dense $\rightarrow$ 1: $E_{LT}$\\Long-Term Energy \\
    \bottomrule
\end{tabular}
\captionof{table}{Architecture for the \textbf{long-term energy function}. The energy computed evaluates the whole trajectory by leveraging 1D convolutional layers with a final temporal average pooling. Number of layers specified in parentheses.}
\label{fig:conv-ebm}
\end{minipage}}
\scalebox{0.81}{
\begin{minipage}{0.4\linewidth}
\vspace{.2cm}
\centering
\begin{tabular}{c}
    \toprule
    \toprule
    Node $\rightarrow$ Edge \\
    \midrule
    5x1 CNN Block Down (2) 64\\ 
    \midrule
    Unfold Trajectory K:5, S:1 \\
    \midrule
    Dense 64\\ 
    \midrule
    MLP Conditioning Block (2) 64 \\
    \midrule
    MLP Conditioning Block  (2) 64 \\
    \midrule
    Edge $\rightarrow$ Node \\
    \midrule
    MLP (2) 64 \\
    \midrule
    Dense $\rightarrow$ 1: $E_{ST}$\\ Short-Term Energy \\
    \bottomrule
\end{tabular}
\captionof{table}{Architecture for the \textbf{short-term energy function}. The energy computed evaluates chunks of 5 steps of a trajectory, obtained with strides of size 1. Number of layers specified in parentheses.}
\label{fig:trans-ebm}
\end{minipage}}
\scalebox{0.8}{
\begin{minipage}{0.4\linewidth}
\vspace{-1.3cm}
\centering
\begin{tabular}{c}
    \toprule
    \toprule
    Node $\rightarrow$ Edge \\
    \midrule
    5x1 CNN Block Down (3) 64\\ 
    \midrule
    Temporal Avg. Pool \\
    \midrule
    MLP (2) 64\\
    \midrule
    Edge $\rightarrow$ Node \\
    \midrule
    MLP (2) 64 \\
    \midrule
    Node $\rightarrow$ Edge \\
    \midrule
    MLP (2) 64 \\
    \midrule
    Dense + LN $\rightarrow$ (L$\times$Num. edges)\\
    \bottomrule
\end{tabular}
\captionof{table}{Architecture for the \textbf{encoder}. Number of layers specified in parentheses.}
\label{fig:encoder}
\end{minipage}}

\subsection{Experimental details}\label{sec:experiment_det}

In the following section we discuss the specific setting for each one of the experiments. In all cases, \model uses Adam optimizer and a learning rate of $LR=3e-4$ with a scheduled decay of $\gamma=0.5$ every 100k iterations.
The illustrated trajectories shown in the figures have been plotted by accumulating the un-normalized velocities of each predicted state to an initial ground-truth coordinate.

\paragraph{Baselines}
We choose the following baselines: 
\begin{itemize}
    \item \textbf{Static}: Copies last input state vector and compares to Ground Truth.
    \item \textbf{LSTM} (multi-node): LSTM model trained to predict the state vector difference at every time-step. It consists of a tow-layer LSTM with shared parameters and 256 hidden units. The input to the model is passsed through a two-layer MLP with ReLU activations before it is passed to the LSTM. Node states are concatenated before being processed by the model. This allows communication across node trajectories. The last hidden vector of the LSTM for each time-steps is also passed through a two-layer MLP with ReLU activations, which outputs a predicted state difference. This is done individually for each particle. The LSTM has access to the ground-truth input states until prediction starts.
    \item \textbf{NRI (learned graph)}: For this model, an encoder infers interactions while simultaneously learning the dynamics from observational data. The encoder outputs a latent code that represents the underlying interaction graph and the reconstruction is based on graph neural networks. For the realistic experiments, we use the LSTM-decoder version, as recommended in their paper. NRI is considered the \textit{de facto} standard for relational inference. The variations of NRI used are: \textbf{(i)}: CNN encoder - MLP decoder for the synthetic experiments and  \textbf{(ii)}: CNN encoder - LSTM decoder for the realistic experiments, as shown in their paper.
    \item \textbf{NRI (full graph)}: This instanciation of NRI is similar to the one above, with the difference that the latent graph is fixed. The encoder is only allowed to output 1 type of edge representation. 
    \item \textbf{dNRI}: Extension of NRI to a dynamic relation setting. The encoder infers separate relation graphs for every time-step. The reconstruction is based on graph neural networks, also in a dynamic fashion. We also use the LSTM decoder version, which produces better results than the MLP decoder version.
    \item \textbf{Conditional GNN}: For this model, we encode the edges similarly as in \model, with the observed part of the trajectory. We decode them in one shot by means of a GNN with message passing.
    \item \textbf{Social LSTM and Directional LSTM (TrajNet++)}: This two models are implemented by \cite{Kothari2020HumanTF} and executed following instructions in their github \url{https://github.com/vita-epfl/trajnetplusplusbaselines}. The architectures are LSTM-based, and use different types pooling functions to gather information surrounding each node. The code is addapted to a state with dimensionality 4 instead of 2.
    \item \textbf{Planning with Diffusion (PwD)}: Consists on a diffusion probabilistic model that plans by iteratively denoising trajectories.
\end{itemize}
 
\paragraph{Quantitative Comparison}
Our setting is the following: we train and evaluate the baselines to predict 20 time-steps in the future after observing a partial trajectory. Predicted states come immediately after the observed and fixed time-steps.

We firstly generate Springs and Charged datasets following the code provided by the authors of \cite{Kipf2018NeuralRI}. We use interaction strengths 5 and 1 respectively. The data is normalized using the statistics of the first 49 states of the training data. \model is trained with 2 energy functions and latent size per edge potential of {$D_z=64$}. We use hidden layers of size 256. We encode 49 time-steps into a set of potentials, fix 1 time-step from the ground-truth and predict the following 20.  We use a number of sampling steps $M=5$ and a step-size of {$\lambda=0.4$}. We use a batch size of 40 and train for 500 epochs. We find it beneficial (although not necessary) to supervise all the $M=5$ trajectories sampled with Langevin, with an exponential weighting scheme. The selected baselines are trained for 500 epochs with the same scheme as \model. Following their indications, we use teacher forcing for NRI, but we supervise predicted states of 20 time-steps. For the LSTM case, the model unrolls 20 time-steps predictions during training, and supervises all outputs of the model, including the burn-in predictions as in \cite{Kipf2018NeuralRI}. For the C-GNN, we encode 50 time-steps into edge conditions of size $D_z=64$ and predict the following 20. In all cases, best models are selected through validation.

Similarly, we run experiments in NBA SportsVU dataset, normalizing with the training data statistics. C-GNN is trained similarly as before with a batch size of 8 for 25 epochs. For prediction in NBA experiments, \model encodes 40 time-steps into a set of potentials of dimension  {$D_z=64$}, fixes 5 time-steps from the ground-truth and predicts the following 20. In training and testing, only the 20 predicted time-steps are generated and supervised with the ground-truth trajectory. We use hidden layers of size 64 due to computational limitations. For LSTM, we use the same training scheme as in the synthetic datasets.
Our model is trained for 25 epochs, with a batch size of 6 and single set of edge potentials. Number of sampling steps $M$ varies from 3 to 5 along the first 200k iterations and a step-size of {$\lambda=0.4$}.

For this dataset, results in social interaction-based networks (S-LSTM, D-LSTM) are very poor in long term. This behavior has been discussed with the authors of D-LSTM (\cite{Kothari2020HumanTF}), and concluded that it is expected when solving a task of prediction. This models are usually employed for tasks such as pedestrian collision avoidance. After switching the objective function to a purely MSE loss and changing the training strategy with the authors' help, the results improved very slightly. The model converged after 25 epochs. 

NRI and dNRI are trained with a batch size of 40, a learning rate of $LR=5e-4$ and hidden sizes of 256. We train them for 50 epochs. For both architectures, we use 2 edge types as recommended in the respective papers. Specifically, in NRI they try for higher number of edge types and conclude that the model is overfitting. For dNRI, the best results both in short and long-term are met by using teacher-forcing in 1 time-step predictions. Results for 20 time-step unrolls are significantly worse.

Finally, experiments in JPL Horizons dataset are with the same baseline setting as NBA SportsVU. In this case, we train all models for 2000 epochs (given the small size of the dataset). In none of the cases we see signs of overfitting. C-GNN and \model use a batch size of 10 and 6 respectively.   \model has a number of sampling steps $M$ that varies from 3 to 5 along the first 300k iterations and a step-size of {$\lambda=0.4$}.
In this experiment NRI and dNRI are trained for 2000 epochs with the same hyperparameters as in the NBA experiments. For both experiments we observe that dNRI performs worse than NRI in long-term. In this case, this is expected as relation-types do not change in time. 

\paragraph{Recombination}
For recombination, we use a mixture of the Springs and Charged dataset, with interaction strengths of 0.1 and 0.5 respectively.
While not necessary, we found recombinations slightly more natural-looking by leveraging a variation of the original architecture. In this case, each EBM has a branch that evaluates trajectory energies unconditionally. That branch is solely used by edges that have been masked out (i.e. conditioned in another EBM). The unconditional architecture is therefore very similar to that in \ref{fig:conv-ebm} and \ref{fig:trans-ebm} but disregarding the conditioning blocks.
We train \model for 120 epochs both in Charged and Springs datasets separately, with a small latent size per potential of {$D_z=8$}. For this experiment, \model is trained to reconstruct 30 observed time-steps and predict the following 10, as we find prediction helpful for proper potential learning. We use a dataset instance with double sampling frequency than for the other experiments. This task is especially challenging, given that we are mixing different data distributions. While experiments can be carried out with the usual setting, we find some additions helpful for better-looking results. i) We perform data augmentation by selecting randomly the initial point of the trajectories in the range $T_0=[1,\dots,50]$, as velocity distributions diverge along the trajectories. ii) The encoder observes the input data both rotated and instance-normalized. iii) We plot the trajectories by fixing the first location and accumulating the unnormalized velocity states. At test-time, we sample $M=6$ times with a step-size of $\lambda=16$. 


\paragraph{Out-of-distribution Detection}
We utilize the same Springs and Charged dataset variations as for the recombination experiments. For evaluation we also utilize the Charged-Springs dataset explained in the main body of the paper. The energy values are obtained at the node level by evaluating a ground-truth trajectory. The hyperparameters of the model are those of the recombination experiments.

For the NBA SportsVU dataset we train a model with the same hyperparameters as for quantitative comparison. However, we discard the node corresponding to the ball. We later evaluate the trained model in the same dataset but including the ball and excluding one of the players. In both cases, we will have 10 agents.

We found especially useful to use the regularization in Equation \ref{eq:cd_loss}. This maintains the in-distribution energy close to 0, while increasing energy from out-of-distribution nodes.

A limitation found is that if a particle with new dynamics behaves like one in-distribution it might have lower energy than a \textit{hard} in-distribution sample. For instance, charged particles at long distances might behave like free-of-charge samples, which is one of the modes of the Springs dataset. Similarly, a ball carried by a player might behave like a player and therefore result in low energy. This can explain why the ball is not detected 30\% of the time. Once again, we illustrate the trajectories by accumulating the un-normalized velocities, which leads to smoother trajectories.

\paragraph{Flexible Generation}
For this experiment, we train a model in NBA SportsVU dataset with the same hyperparameters as for quantitative comparison. However, in this case we only reconstruct 40 time-steps. 
For the Charged dataset experiments, we also utilize the same setting as in training. The illustrations are made by accumulating the un-normalized velocities. We skip the initial ground-truth timesteps and plot only the generated trajectories.
The formulation of the hand-crafted potentials is described with detail in the main body of the paper. $\lambda$ corresponding to the weight of the new potential is found by grid search by means of visual inspection. However, it is fixed across instances of the experiment.
Extreme values for $\lambda$ will yield unrealistic results. 

\newpage
 \subsection{Complexity Analysis} \label{sec:comp-analysis}

\begin{wrapfigure}{R}{0.4\textwidth}
\vspace{-12pt}
\small
\centering
\begin{tabular}{l c c }
\toprule
 \makecell{Time (s) \\ per iteration \\ and sample}	& Train	& Test \\
\midrule
NRI (CNN+MLP)       &1.96e-3 & 7.06e-4     \\
NRI (CNN+MLP)       &4.96e-2	&4.33e-2    \\
\model (3 steps)      &3.91e-3	&1.78e-3   \\
\model (6 steps)      &5.91e-3	&3.40e-3    \\
\bottomrule
\end{tabular}
\captionof{table}{\small \textbf{Time Complexity Analysis of \model w.r.t an auto-regressive baseline}. \model has similar time complexity than NRI. The overhead due to the iterative generative process is compensated with the one-shot nature of our predictions. }
\label{tab:time-complexity}
\vspace{-12pt}
\end{wrapfigure}

Table \ref{tab:time-complexity} shows a temporal complexity analysis of NRI, an auto-regressive baseline, and our model. We compare two versions of NRI: One with an MLP decoder (synthetic datasets) and one with a RNN decoder (NBA, Horizons). Both have a CNN encoder which is fairly similar to ours. As \model computation time varies with the number of sampling steps, we provide two measurements at 3 and 6 sampling steps for generation. We find that the computation time of \model is similar to that of baselines.

The EBM decoder of \model generates samples by computing gradients which takes approximately $\times 2$ longer than a feedforward network, as it must compute the gradients for the backward pass. To generate a single sample, \model further computes $N$ Langevin steps. In our experiments, $N$ ranges from 3 to 6. This is an additional factor of $\times N$. However, \model predicts trajectories in one-shot. This differs from baselines such as NRI, where states are predicted auto-regressively. The length $T$ of the generated sequence doesn't have a significant impact on complexity of \model, while it adds a factor of $\times T$ to auto-regressive approaches. Thus, this results in a similar time complexity for auto-regressive baselines and \model.

 \subsection{More Examples and Additional Results} \label{sec:add_results}
 
In this Section, we illustrate more examples of the main experiments and additional quantitative results.

\paragraph{Downstream tasks: Edge-type Prediction}

\begin{wrapfigure}{R}{0.35\textwidth}
\vspace{-12pt}
\centering
\small
\begin{tabular}{l c c }
\toprule
 & jointLSTM & \model (Ours)   \\
\midrule
 Springs        & 99.3\% & \textbf{99.9}\%     \\
Charged         & 58.9\% & \textbf{69.4}\%    \\
\bottomrule
\end{tabular}
\captionof{table}{\small \textbf{Edge Classification Accuracy}. We evaluate the \model's learned representations. Given our learned latent codes of size 64, we train a linear layer to classify the edge types into their ground-truth types. }
\label{tab:downstream}
\end{wrapfigure}

Following we provide the results of \model in the downstream task of edge classification, given the learned representations. We do this for Springs and Charged datasets, where we do have ground-truth edge types. Given our learned representations of size 64, we train a linear layer to classify the edge types into their ground-truth types. We obtain the accuracy in \ref{tab:downstream}

As we can see, \model is better than the baseline in both settings. In the Charged dataset, the accuracy is substantially lower than for the Springs dataset. A plausible explanation is that particles at long distances have close to negligible pairwise attracting or repelling forces. This effect is not as present in the Springs dataset.

\paragraph{JHMDB Experiments}

\begin{wrapfigure}{R}{0.35\textwidth}
\small
\centering
\vspace{-10pt}
\begin{tabular}{l|cc}
\toprule
& \multicolumn{2}{c}{\textbf{JHMDB}}  \\ \midrule
Prediction steps & 1 & 10  \\ \midrule
NRI & 6.71e-4	& 8.98e-3 \\
dNRI & 9.21e-4	&  1.32e-2\\
\model (Ours) & \textbf{3.05e-4} &	\textbf{3.41e-3} \\
\bottomrule
\end{tabular}
\caption{\small \textbf{Mean squared error (MSE) in predicting future states for JHMDB dataset}, with 12 nodes for each limb joint. \model performs better than the baselines.}\label{tab:pred_mse_nba_jpl}
\vspace{-10pt}
\end{wrapfigure}

We include a real-world trajectory forecasting experiment in the Joint-annotated Human Motion Data Base \cite{jhmdb}. It contains 2D skeleton trajectories from 36-55 clips per action class with each clip containing 15-40 frames. There are 31,838 annotated frames in total, from which we filter the ones that contain at least 30 timesteps, keep the 12 joints associated to the limbs and normalize in the range [-1,1]. This results in ~1000 training trajectories, ~100 for validation and 215 for testing. We observe 16 timesteps, fix 4 and predict the following 10. We compare to the NRI baseline with an LSTM decoder, and dNRI baseline. NRI will observe the 20 first timesteps and predict the following 10. Once again, dNRI does not display gains with respect to NRI. 

\newpage
\paragraph{Qualitative Examples}

\begin{wrapfigure}{l}{0.6\textwidth}
\centering
  \includegraphics[ trim={50 320 915 0}, width=0.6\textwidth]{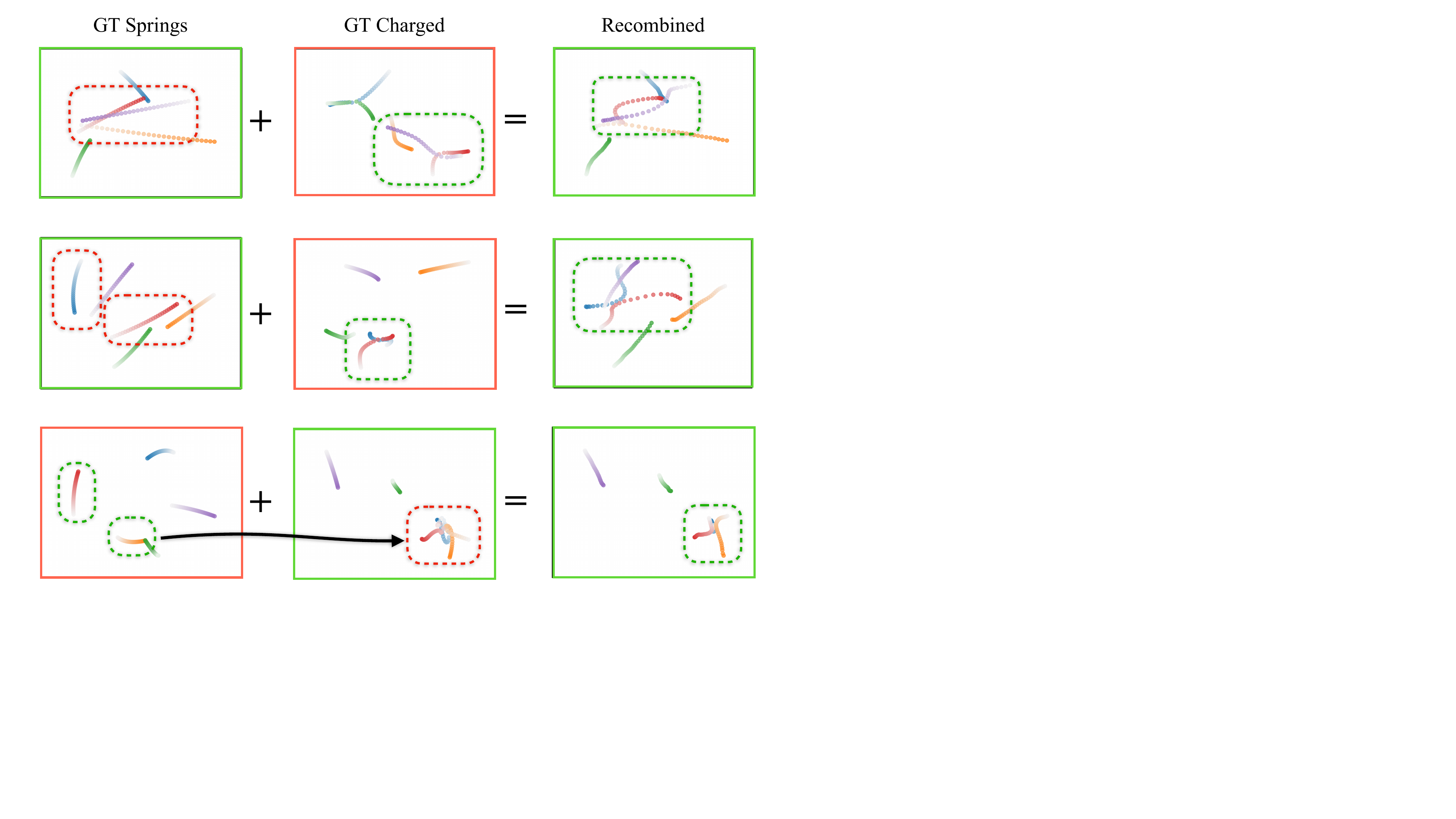}
  \caption{\small \textbf{More examples of recombinations.} Illustrated, samples from Springs (Col. 1) and Charged (Col. 2) and their recombinations (Col. 3). \model encodes both trajectories. \model is able to reconstruct trajectories framed in green with the initial conditions marked in blue, while swapping the potentials associated to the nodes in the red dashed box. Recombinations look semantically plausible and smooth.}
\label{fig:add_recomb}
\vspace{-8pt}
\end{wrapfigure}

Figure \ref{fig:add_recomb} illustrates cross-dataset combinations of potentials. We encode ground-truth trajectories  (columns 1 and 2) separately into their corresponding potentials. We train a different model for each data distribution. Next, we fix initial conditions $\V{x}_0$ (blue points) of the scene framed in green and aim to reconstruct them. In test-time we generate a trajectory by minimizing simultaneously the energy functions corresponding to the two models. Each model targets specific edges. Particularly, one model adds potentials corresponding to the mutual edges of the particles highlighted by a red dashed box, while the other aims to reconstruct the rest of the trajectories. 

\begin{figure*}[h]
\centering

  \includegraphics[trim={10 340 410 0}, width=\textwidth]{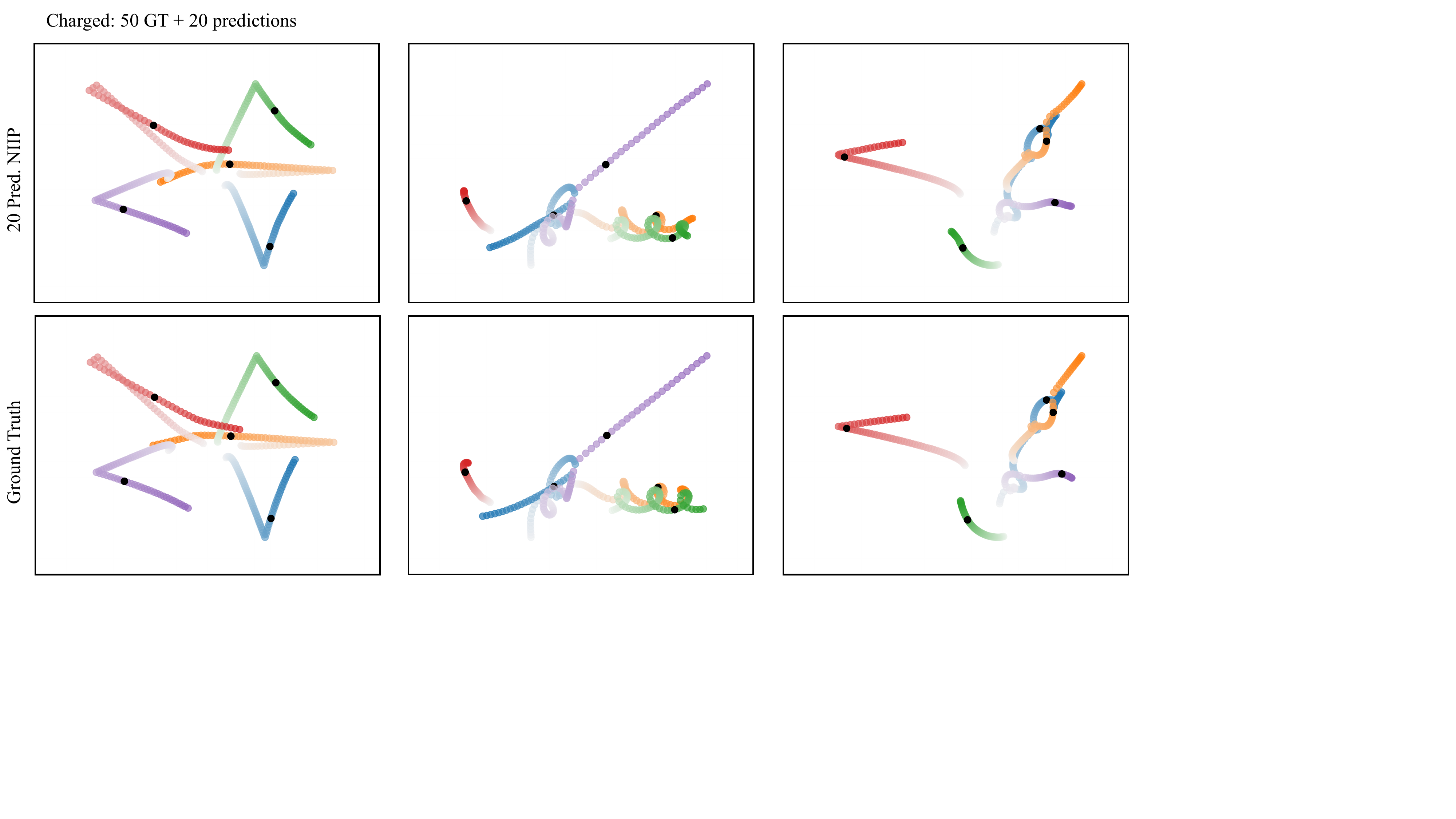}
  \caption{\small \textbf{Qualitative example of 20 predictions in the future of \model in the Charged particles dataset.} In both rows, first 50 steps are the ground-truth. The black dot indicates the beginning of predictions for row 1. In all cases, predictions are fairly close to the ground-truth. In the cases where they differ (green node - center) the predictions are smooth and look reasonable.}
\label{fig:pred_charged}
\end{figure*}

Next, Figure \ref{fig:pred_charged} show qualitatively the predictions of \model compared to the ground truth in the Charged dataset. The initial 49 steps are the ground-truth trajectory. The black dot indicates the beginning of our model's predictions (the trajectory color gets lighter with time). Predictions are often accurate. In cases where there is a significant difference with the ground-truth (e.g. green node in the center of Figure \ref{fig:pred_charged}), the predicted trajectory looks semantically plausible.

\begin{figure*}[b]
\centering
  \includegraphics[trim={30 360 540 0}, width=\textwidth]{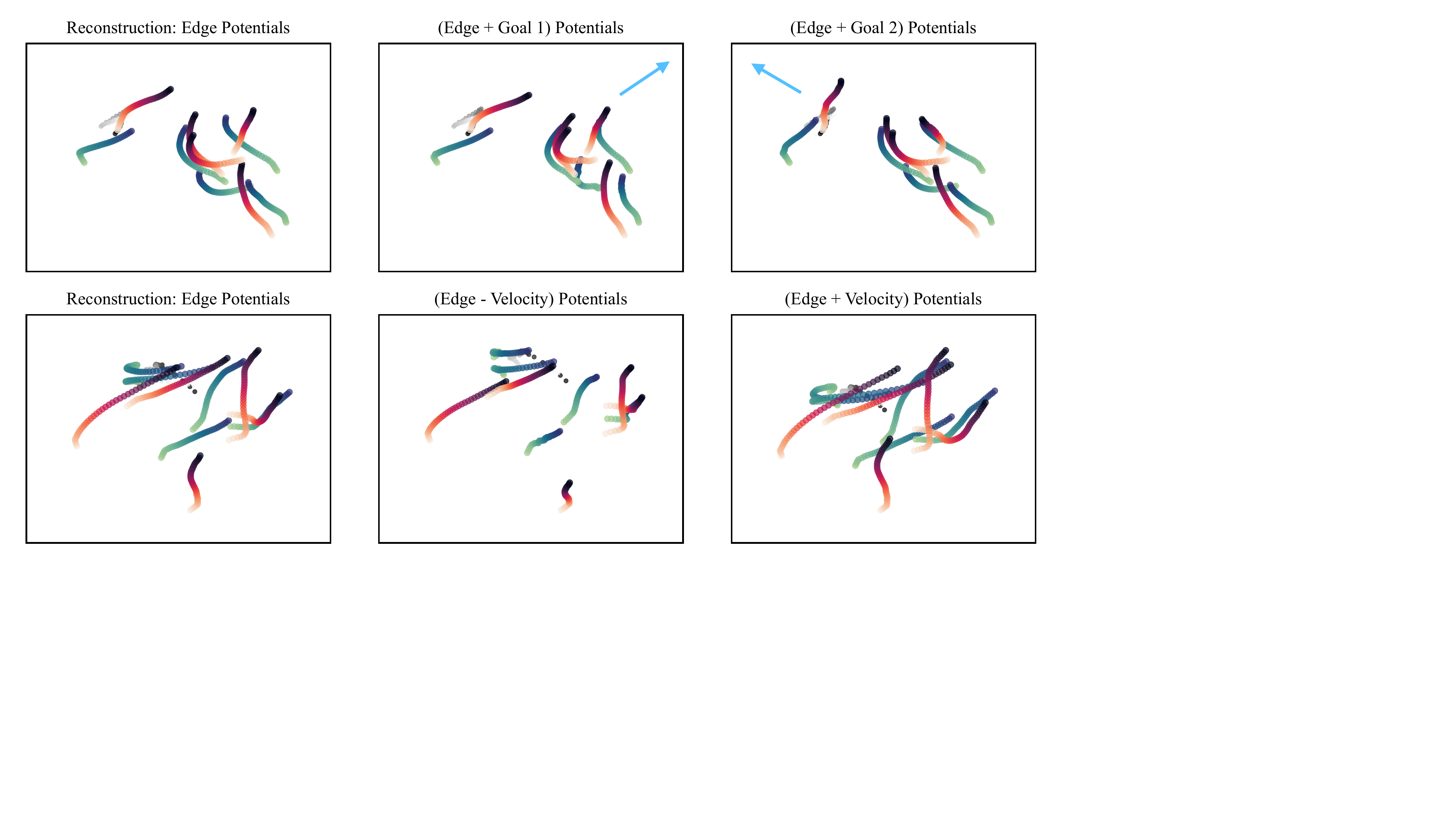}
  \caption{\small \textbf{More examples of NBA reconstructed trajectories with added potentials in test-time.} \model can generate realistic trajectories by respecting both \textbf{(i).} The learned edge potentials + \textbf{(ii).} new hand-crafted potentials added in test-time. Illustrated (top row) we see the reconstruction of the trajectory with an added goal. We can see that the trajectories have the tendency to be attracted in the direction indicated by the blue arrow. The actual goal is located outside the frame. We can see (bottom row) potentials of higher and lower velocity than the reconstruction. In all cases, the trajectories follow the new potentials while respecting the original trajectory shapes.}
\label{fig:add_new_constraints}
\end{figure*}

Figure \ref{fig:add_new_constraints} illustrates the ability of \model to add hand-crafted potentials in test-time. In the top row, a trajectory reconstruction together with added goals that the nodes (players) are attracted to. In the bottom row, a trajectory reconstruction followed by the same trajectory with variations of the agent velocities. The resulting plots show how both the edge potentials and the newly added hand-crafted potentials are respected.

\begin{figure*}[h]
\centering
  \includegraphics[trim={20 40 660 0}, width=\textwidth]{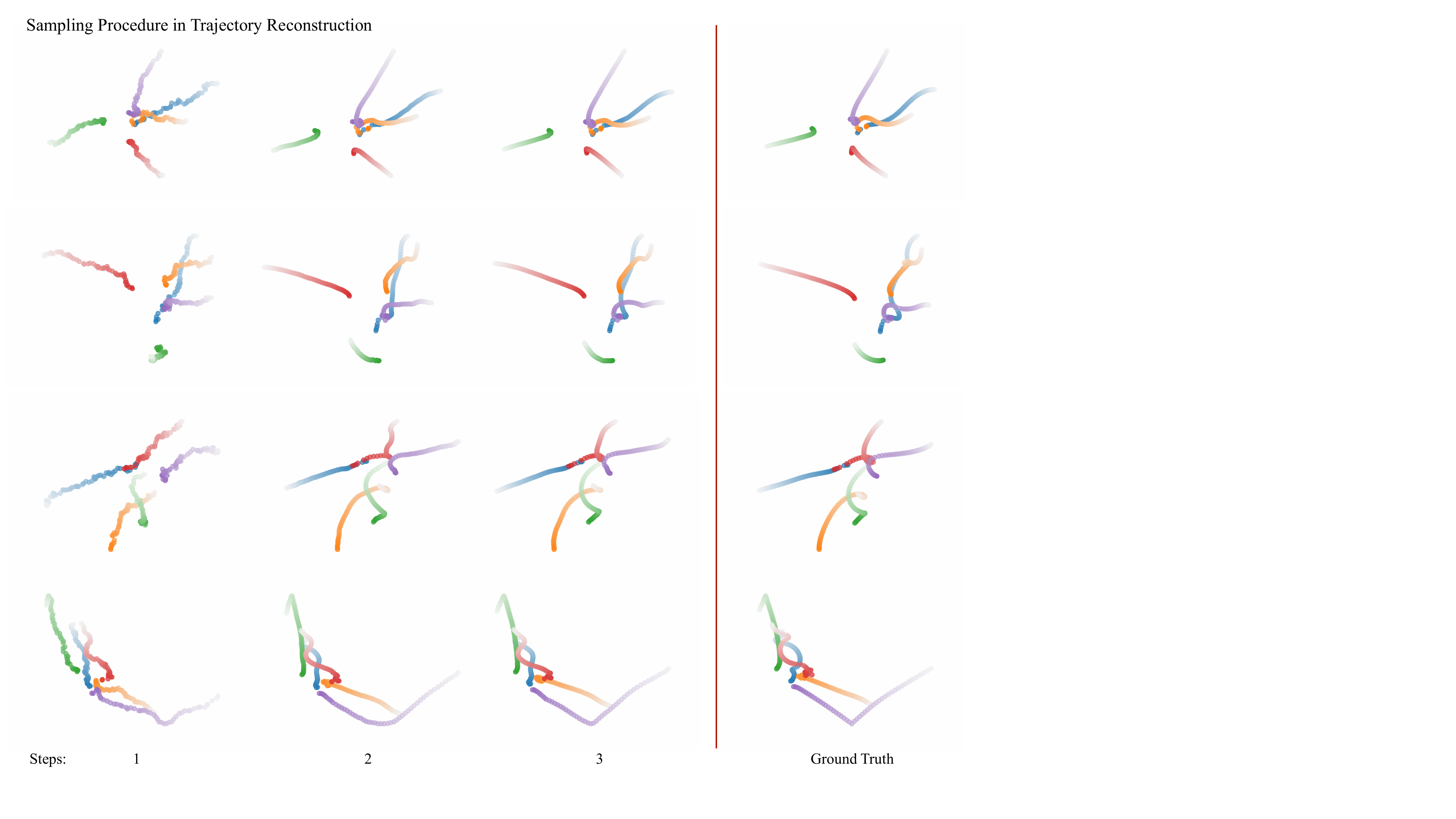}
  \caption{\small \textbf{Examples of trajectory reconstruction procedure for 50 time-steps of the Charged dataset}. We initialize the velocity as uniform noise and optimize with 3 Langevin Dynamics steps. We obtain a faithful reconstruction of the ground-truth trajectories.}
\label{fig:sampling}
\end{figure*}

Finally, Figure \ref{fig:sampling} illustrates an example of the sampling procedure. By leveraging Langevin Dynamics sampling we refine our predictions iteratively in a gradient-based optimization procedure. The model quickly understands the general shape of the trajectories and refines them locally using the final sampling steps.

\subsection{{Ablation Study}} \label{sec:ablation}

{Following, we add an ablation study carried on in the Charged dataset, with 150 epochs and a batch size of 40. We we utilize also smaller filter size of 64 and we supervise only the last step of Langevin sampling, yielding sub-optimal performance. However, it allows us to ablate the different options of our approach. Here, we analyse \model's performance through a variety of design choices. Those are the following:
\begin{itemize}
    \item Latent size: We explore different sizes for the edge potentials. The choices are LS $\in \{16, 32, 64, 128\}$. 
    \item Langevin step size. The choices are $\lambda \in \{0.1, 0.2, 2.0, 6.0, 10.0, 14.0\}$. 
    \item Objective: We evaluate the impact of adding the contrastive divergence objective to the MSE reconstruction loss.
    \item Edge masking strategy: In a setting with 2 EBMs, we evaluate 1) a random masking strategy, 2) a masking strategy based on the edge contribution to a specific node and 3) no masking strategy.
    \item Decoder baseline: We substitute the iterative energy minimization by a feed-forward graph decoder. This is equivalent to the Conditional GNN baseline.
\end{itemize}
For this experiment, we follow the setting in the quantitative comparison and simply modify the variable of interest. That analysis shows how a small langevin step is desirable in terms of performance ($\lambda = 0.2$). The model seems to be robust to the latent size choice, although there is a slight preference for $LS=32$. When it comes to the objective, we show quantitatively that regularizing training with a contrastive divergence term improves performance. We can also observe how a masking strategy is better than none. Finally, we show by comparing to a decoder baseline how our sampling strategy is crucial for competitive results. Our ablation study is summarized in Table \ref{tab:ablation_study}.}

Similarly, we perform an ablation at test-time, with our best trained model for the Charge dataset. We analyse \model's performance under the following variations:
\begin{itemize}
    \item Number of Langevin steps: We train our model with $M=6$ langevin steps and test it with $M \in \{1,2,3,4,5,6,7,8,9,10,20\}$.
    \item Number of nodes: We train \model with 5 nodes and test it in datasets with $N=3$ and $N=7$
\end{itemize}
Results are summarized in Table \ref{tab:ablation_test}. "Train" indicates the setting used in training. In this case we train for 300 epochs with a batch size of 40.

\begin{figure}[h] 
\centering
\small
\begin{tabular}{l c c c}
\toprule
\textbf{Ablation Study} & \textbf{1 step} & \textbf{10 step} & \textbf{20 step}   \\
\midrule
Latent Size\\
\midrule

16            & 1.20e-3    & 3.75e-3  & 7.06e-3 \\
 32             & 9.87e-4    & 3.60e-3  & 6.82e-3 \\
 64             & 1.15e-3    & 3.82e-3  & 7.00e-3 \\
 128            & 1.15e-3    & 3.91e-3  & 7.24e-3\\
 
 \midrule
Sampling step size\\
\midrule
0.1               & 9.54e-4    & 3.53e-3  & 6.81e-3  \\
0.2                & 9.07e-4    & 3.36e-3  & 6.62e-3  \\ 
2.0                &1.16e-3    &3.82e-3  & 7.09e-3  \\
6.0                & 1.15e-3    & 3.82e-3  & 7.00e-3  \\
 10.0                & 1.14e-3    & 3.78e-3  & 7.07e-3  \\
 14.0                & 1.19e-3    & 3.91e-3  & 7.29e-3  \\
 \midrule
Masking type\\
\midrule
 Mask random           & 1.15e-3    & 3.82e-3  & 7.02e-3  \\
 Mask by node          & 1.15e-3    & 3.78e-3  & 7.03e-3  \\
 No masking            & 1.17e-3    & 3.82e-3  & 7.09e-3  \\
\midrule
Objective\\
\midrule
 MSE+CD                & 1.15e-3    & 3.82e-3  & 7.02e-3  \\
 MSE (no CD)           & 1.36e-3    & 4.02e-3  & 7.47e-3  \\
\midrule

\midrule
 Decoder (no sampling) & 3.27e-3    & 5.31e-3  & 9.65e-3  \\
\bottomrule
\end{tabular}

\captionof{table}{\small \textbf{Ablation study} investigating effects of different factors like latent size, sampling step size, masking type and objective in results.}
\label{tab:ablation_study}
\end{figure}

\begin{figure}[h] 
\small
\centering
\begin{tabular}{l c c c}
\toprule
\textbf{Ablation Study in Test} & \textbf{1 step} & \textbf{10 step} & \textbf{20 step}   \\
\midrule
Number of Langevin steps $M$\\
\midrule

1          & 1.06e-1     & 1.18e-1     & 2.47e-1   \\
2           & 3.23e-2     & 4.34e-2     & 8.22e-2   \\
 3           & 7.00e-3     & 1.16e-2     & 2.68e-2    \\
 4           & 1.86e-3     & 4.63e-3     & 1.00e-2    \\
 5            & 1.03e-3     & 3.51e-3     & 7.05e-3     \\
 6 (Train)            & \textbf{8.85e-4 }& \textbf{3.33e-3} & 6.54e-3    \\
 7           & 9.00e-4     & \textbf{3.33e-3} & \textbf{6.45e-3}  \\
 8            & 8.93e-4     & 3.36e-3     & 6.47e-3     \\
 9          & 9.00e-4     & 3.39e-3     & 6.60e-3    \\
 10          & 9.10e-4     & 3.44e-3     & 6.66e-3     \\
 20          & 1.03e-3     & 3.64e-3     & 7.04e-3   \\
 \midrule
Number of nodes $N$\\
\midrule
3   & 1.79e-3     & 3.71e-3     & 6.74e-3             \\
5 (Train)    & \textbf{8.85e-4} & \textbf{3.33e-3} & \textbf{6.54e-3}  \\
7       & 1.34e-3     & 4.66e-3     & 1.01e-2        \\
\bottomrule
\end{tabular}

\captionof{table}{\small \textbf{Ablation study} investigating effects of different factors in test-time: Number of langevin steps and number of nodes.}
\label{tab:ablation_test}
\end{figure}

\subsection{Broader Impact} \label{sec:broad}
Understanding interactions across agents in a trajectory is fundamental to explain their present behavior and predict their future. The importance of such understanding is higher when we do not have access to the true interaction types or they are simply not a discrete set. In those cases, being able to learn representations of interactions from observational data provides a window into the physics of the world we live in. These property is desirable in AI for applications such as molecular dynamics modeling or autonomous vehicles, which have a huge impact on our lives. Despite the fact that there is a wide range of approaches for inferring interactions and predicting trajectories, there is relatively little work on inferring these in a interpretable and compositional manner. Our model aims to learn interaction potentials that allow for a higher degree of manipulation over the learned representations. This interpretability and manipulability properties are important to AI, but might raise concerns of abuse. Our approach, similar to many other approaches,  may capture the implicit biases present in data. There is also the potential threat of attacks to systems that rely on interpretable models, which can be more easily targeted than those which are opaque.

\end{document}